\documentclass{article}

 \usepackage[final]{neurips_mlsb_2024}

\usepackage[utf8]{inputenc} %
\usepackage[T1]{fontenc}    %
\usepackage{hyperref}       %
\usepackage{url}            %
\usepackage{booktabs}       %
\usepackage{amsfonts}       %
\usepackage{nicefrac}       %
\usepackage{microtype}      %
\usepackage{xcolor}         %
\usepackage{amsmath}
\usepackage{graphicx}
\usepackage{mathtools}
\usepackage{wrapfig}
\usepackage{booktabs}
\usepackage{caption}
\title{PairSAE: Mechanistic Interpretability from  Pair Representations in Protein Co-Folding}

\author{%
 Giosue Migliorini\thanks{Work done during an internship at Flagship Pioneering.} \\
  University of California, Irvine\\
  \And
 Aristofanis Rontogiannis \\
  Flagship Pioneering\\
  \And
 Grigori Guitchounts \\
  Flagship Pioneering\\
  \And
 Nicholas Franklin \\
  Flagship Pioneering\\
  \And
 Axel Elaldi \\
  Flagship Pioneering\\
  \And
 Olivia Viessmann \\
  Flagship Pioneering\\
}

\begin{document}

\maketitle

\begin{abstract}
Foundation models for structural biology have achieved remarkable performance in predicting biomolecular structure and show promise for the design of proteins and small molecules. Yet understanding \emph{which} internal features drive their outputs remains challenging. Standard sparse autoencoders (SAEs), effective on transformer-style \emph{sequence embeddings}, do not transfer cleanly to pairformer-like architectures: naïvely operating on pairwise representations yields a quadratic blow-up of features and obscures concepts distributed jointly across sequence and pair representations.
We introduce \textit{PairSAE}, which summarizes pairwise tensors via an $N$-mode SVD into token-wise interaction roles, then uses a sparse autoencoder to learn a \emph{shared} set of token-level features that decode into both sequence and pair representations. Evaluated on Boltz-2 activations for PLINDER protein–ligand complexes, PairSAE yields interpretable features that align with UniProt annotations and predict Boltz-2 affinity values. These results indicate that PairSAE links the latent space of foundation models for structural biology to interpretable structural concepts, clarifying what the model “knows” while avoiding pairformer-induced pitfalls that limit conventional SAEs.
\end{abstract}

\section{Introduction}

Recent advances in protein structure prediction, most notably AlphaFold and the Boltz models \citep{jumper2021highly, abramson2024accurate, wohlwend2025boltz, passaro2025boltz}, have transformed structural biology and now extend to DNA, RNA, small molecules, and their complexes. These models achieve high accuracy without explicitly encoding physical or chemical laws, instead exploiting statistical regularities in large datasets of experimentally determined structures. This raises a central question: to what extent do they implicitly capture the physics and chemistry that govern folding? Answering this requires interpretability -- explanations that let researchers assess prediction reliability and biophysical plausibility -- yet the highly nonlinear architectures of deep models make such analysis challenging. Work in mechanistic interpretability shows that apparent “polysemantic” neurons can arise when networks represent more sparse features than available dimensions, forcing features into superposition rather than cleanly localizing them \citep{elhage2022toy, scherlis2022polysemanticity}. Sparse autoencoders (SAEs) address this by disentangling superposed features into more monosemantic components via sparse, overcomplete dictionaries \citep{olshausen1997sparse, ng2011sparse, elhage2022toy}. Results in large language models \citep{cunningham2023sparseautoencodershighlyinterpretable, bricken2023monosemanticity} and protein language models (pLMs) \citep{simon2024interplm, adamsmechanistic} suggest that SAEs can recover interpretable latent features, making them a suitable candidate for probing the representations learned by protein structure prediction models

However, applying SAEs naively to structure prediction models is nontrivial. 
Unlike most transformer-based LLMs and pLMs, these models use pairformer blocks with pairwise representations alongside sequence embeddings \citep{abramson2024accurate}. Learning separate dictionaries per pair scales quadratically and obscures analysis, and repeated pair–sequence interactions suggest features may be superposed across both spaces.
To address this, we introduce \textit{PairSAE}, which reconstructs both sequence-level and pairwise embeddings from a shared feature set. Applied to Boltz-2, PairSAE links latent representations to interpretable structural concepts, clarifying what the model “knows” about folding and enabling more principled evaluation and discovery in structural biology.

\begin{figure}
    \centering
    \includegraphics[width=0.36\linewidth]{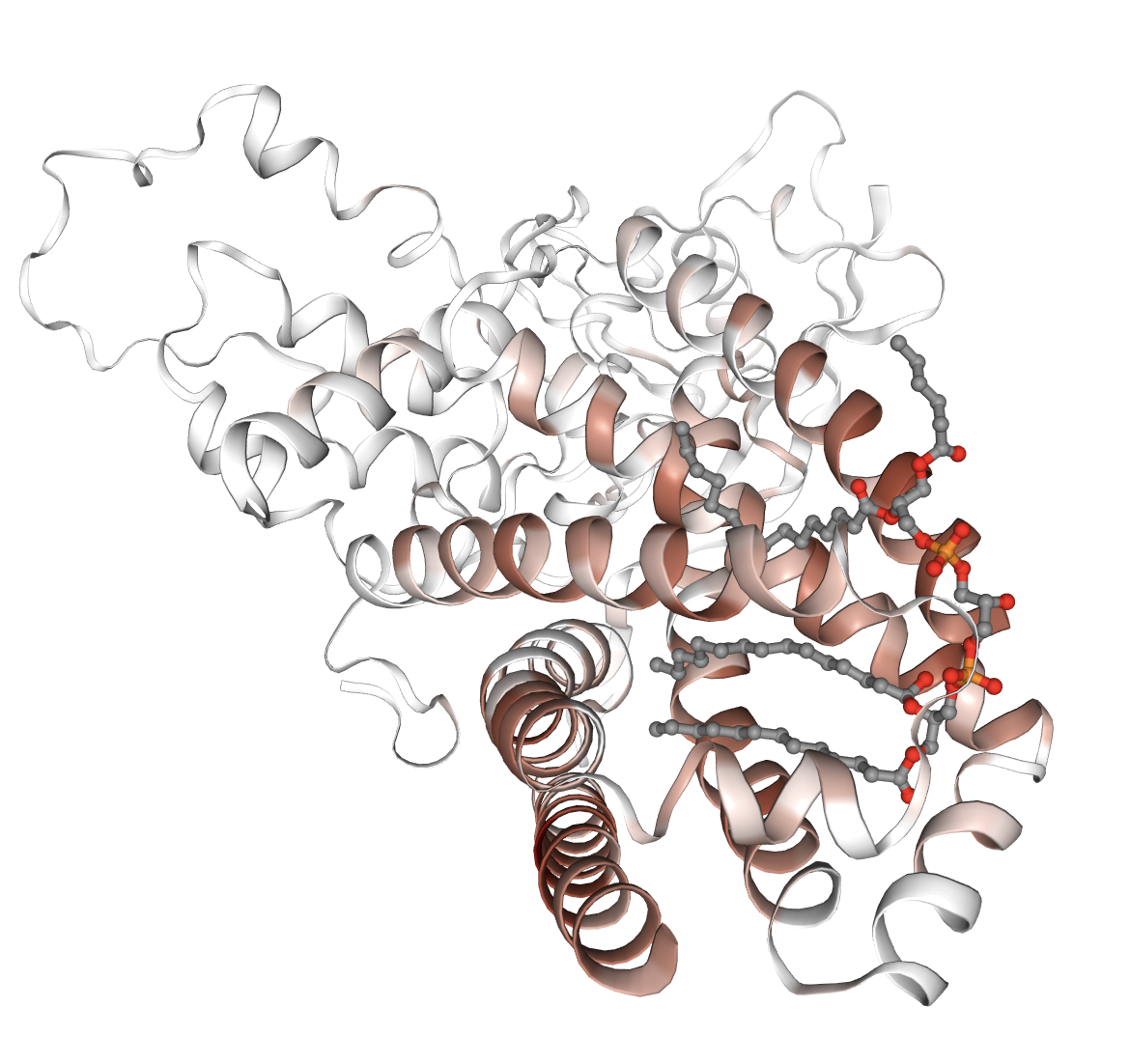}
    \includegraphics[width=0.36\linewidth]{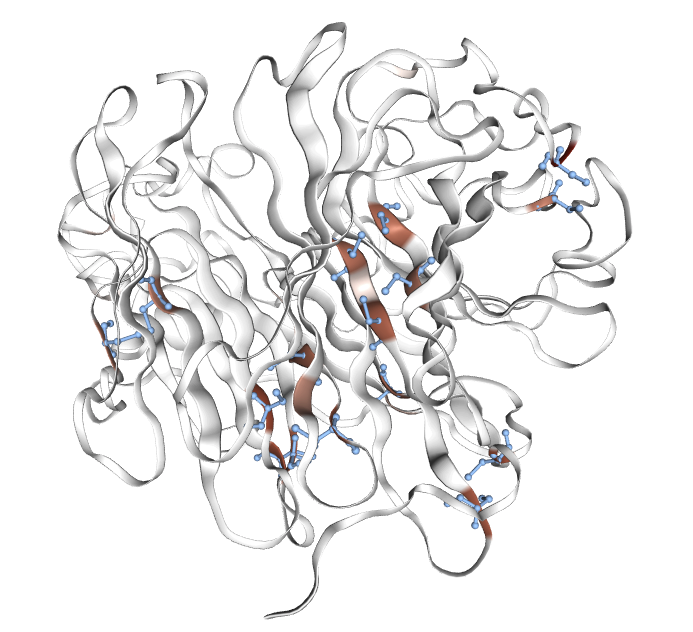}
    \includegraphics[width=0.25\linewidth]{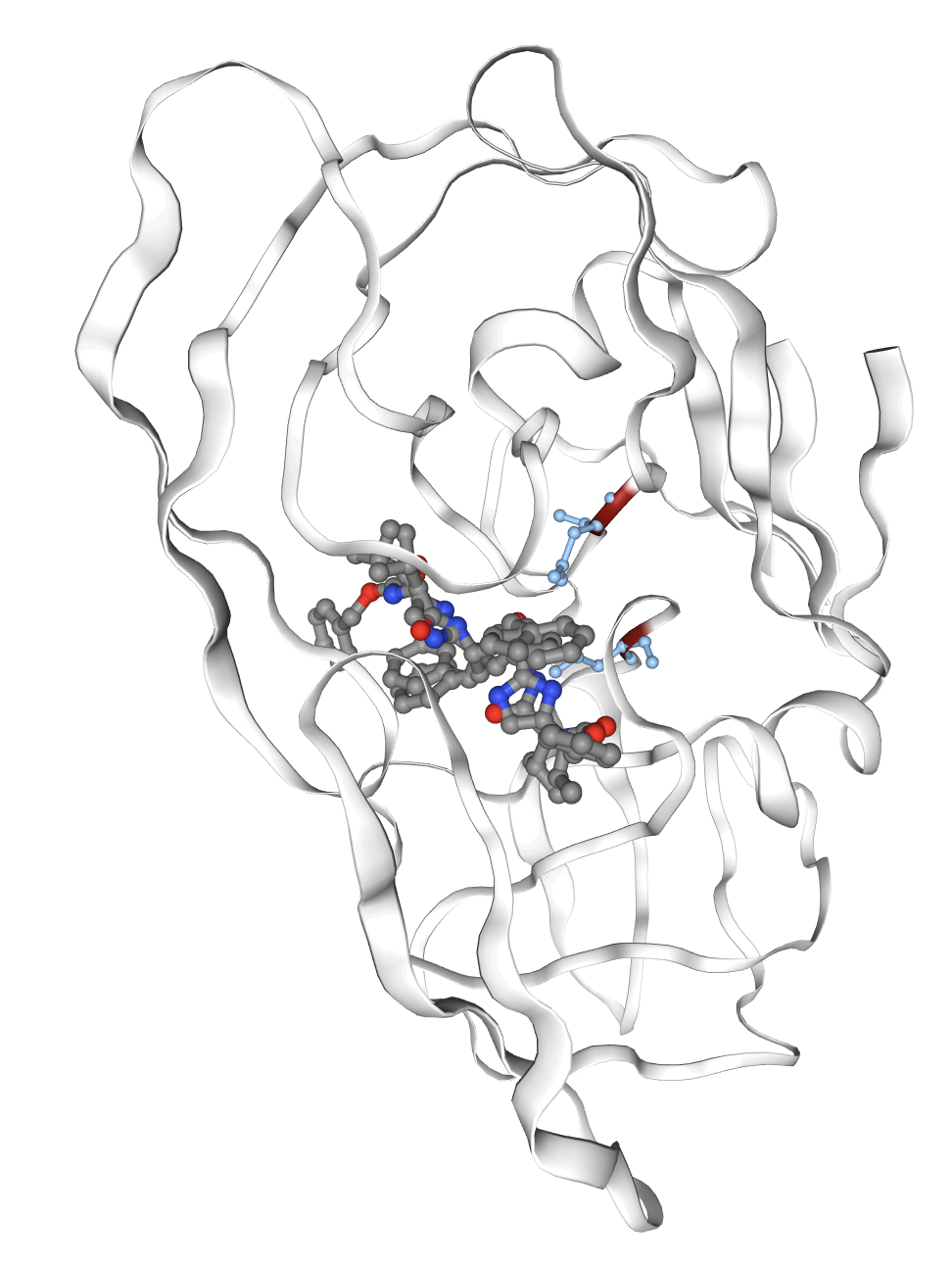}
    \includegraphics[width=0.9\linewidth]{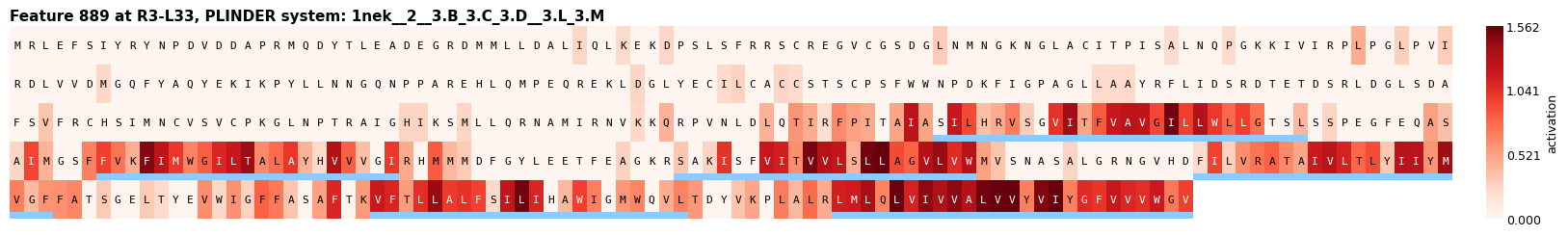}
    \includegraphics[width=0.9\linewidth]{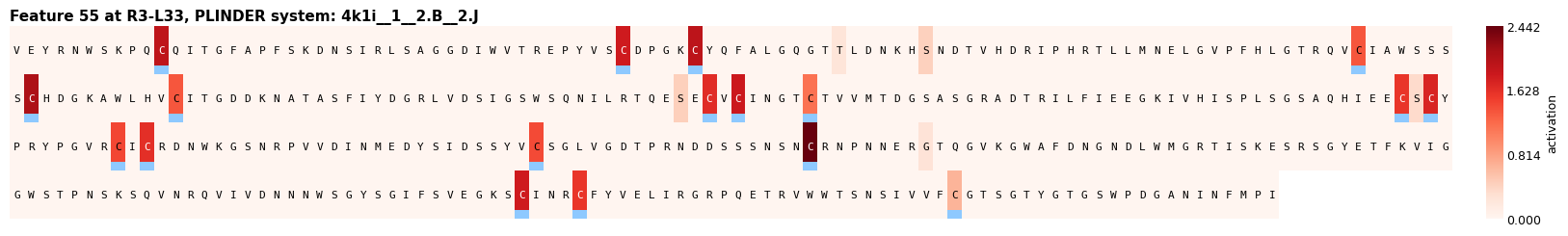}
    \includegraphics[width=0.9\linewidth]{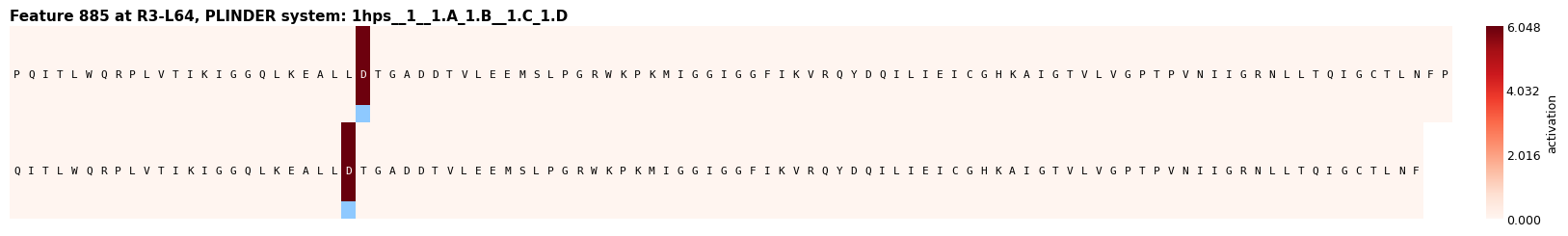}
    \caption{
    Interpretable features recovered by PairSAEs trained at layers 33 and 64 (third recycling step) of Boltz-2. \textbf{Top left:} feature 889 on \textit{E.~coli} Complex II (PDB 1NEK); activates on \textit{\textbf{transmembrane}} proteins (token $F_1=0.58$, complex $F_1=0.65$). \textbf{Top center:} feature 55 on influenza N2 neuraminidase chain B (PDB 4K1I); activates on \textit{\textbf{disulfide bonds}} (token $F_1=0.79$, complex $F_1=0.81$). \textbf{Top right:} feature 885 on an HIV-1 protease inhibitor (PDB 1HPS); activates on \textit{\textbf{protease active sites}} shared with a dimeric partner (token $F_1=0.93$, complex $F_1=0.93$). \textbf{Bottom: }per-residue activations with UniProt ground-truth (blue).
    }
    \label{fig:placeholder}
    \vspace{-0.8\baselineskip}          
\end{figure}

\section{Background}

\subsection{Sparse autoencoders}
Sparse dictionary learning \citep{olshausen1997sparse, lee2006efficient, ng2011sparse} is a long-standing representation learning technique that aims at finding a dictionary $\mathbf{D}=\{\mathbf{d}_j\}_{j=1}^D$ approximating input vectors $\mathbf{x}\in \mathbb{R}^n$ via a linear combination of its columns with a set of sparse latent features $\mathbf{h}\in\mathbb{R}^D$, extracted from $\mathbf{x}$ itself. It has recently been shown that this formulation can yield interpretable features in transformed-based models such as large language models \citep{elhage2022toy, templeton2024scaling, cunningham2023sparseautoencodershighlyinterpretable}, protein language models like ESM2 \citep{simon2024interplm, garcia2025interpreting, adamsmechanistic, gujral2025sparse, parsan2025towards}, and foundation models for genomics like Evo 2 \citep{brixi2025genome}.

Such features can be obtained by training a sparse autoencoder using a reconstruction loss with a sparsity constraint. In mech interp the autoencoder is typically linear, with both a sparsity and positivity constraint on the latent features. A common construction is
\begin{equation}
    \mathbf{h} = \sigma \left( \mathbf{E}\, \mathbf{x} + \mathbf{b}_\text{enc} \right),\quad 
    \hat{\mathbf{x}} \ =\  \mathbf{D}\, \mathbf{h} \ +\ \mathbf{b}_\text{dec},
\end{equation}
where $\mathbf{E}\in \mathbb{R}^{D\times n}, \mathbf{b}_\text{enc}\in \mathbb{R}^{D} 
$ are the weights and bias terms for the encoder, 
$\mathbf{D}\in \mathbb{R}^{n\times D}, \mathbf{b}_\text{dec}\in \mathbb{R}^{n}$ are the weights and bias terms for the decoder, and $\sigma$ is a sparsity-inducing nonlinearity such as ReLU \citep{cunningham2023sparseautoencodershighlyinterpretable, bricken2023monosemanticity}, TopK \citep{gaoscaling}, BatchTopK \citep{bussmann2024batchtopk}, and others.

\subsection{Pairformer representations}
In a pairformer module \citep{abramson2024accurate}, an input sequence $\mathbf{x}$ of length $N_\text{tok}$ tokens is encoded into a 
sequence embedding matrix $\mathbf{S}\in\mathbb{R}^{N_\text{tok} \times n_s}$, whose $i$-th row is the token embedding vector $\mathbf{s}_i \in \mathbb{R}^{n_s}$, and a three-dimensional pair representation tensor $\mathcal{Z}\in \mathbb{R}^{N_\text{tok} \times N_\text{tok} \times n_z}$, whose $(i,j)$-th slice $\mathcal{Z}_{i,j} \in \mathbb{R}^{n_z}$ represents the embedding of token pair $(i,j)$. Here, $n_s$ and $n_z$ are the sequence-level and pairwise embedding dimensions, respectively. Pair representations control the information flow in the updates of the sequence-level embeddings at each layer of the pairformer, by biasing the attention logits \cite{abramson2024accurate}. Pair representations were first introduced with the Evoformer, part of the architecture of AlphaFold 2 \cite{jumper2021highly}.

\section{PairSAE}

\paragraph{Motivation.}
Our goal is to obtain token-level sparse features that jointly reconstruct sequence and pair representations. Pairwise embeddings impose no structure, so we first compress them into a token-wise summary that preserves row/column interaction roles. A shared feature set then decodes into both representation types.

\paragraph{Summarizing pairwise interactions.} We propose to perform an $N-$mode singular value decomposition (SVD) of the tensor $\mathcal{Z}$, and use it to construct a sequence representation that is specifically designed to incorporate information about how each token interacts in the sequence. We include a brief introduction to $N-$mode SVD in Appendix \ref{appdx:svd}.
Information about the role that each token plays \emph{as a row and as a column} in $\mathcal{Z}$ can be obtained by inspecting the mode-1 and mode-2 matrices of left singular vectors $\mathbf{U}^{(1)},\mathbf{U}^{(2)} \in\mathbb{R}^{N_\text{tok}\times N_\text{tok}}$ \citep{de2000multilinear, vasilescu2002multilinear}. We consider a truncation to the first $r$ columns
\(\mathbf{U}^{(1)}_{:,1:r}\) and \(\mathbf{U}^{(2)}_{:,1:r}\), ordered by their singular value. We then concatenate them column-wise into a new sequence level embedding 
\begin{equation}
    \mathbf{m}_i = \big[\mathbf{U}^{(1)}_{i,1:r}\;\|\;\mathbf{U}^{(2)}_{i,1:r}\big], \quad i=1,\dots,N_\text{tok}.
\end{equation} 

\paragraph{Autoencoder.} PairSAE latent features are obtained by encoding a concatenation of this new embedding matrix with the original sequence-level representations from the pairformer model:
\begin{equation}
   \mathbf{h}_i = \sigma \left( \mathbf{E}\,[\mathbf{s}_i\;\|\;\mathbf{m}_i] + \mathbf{b}_\text{enc} \right), , \quad i=1,\dots,N_\text{tok},
\end{equation}
where $\mathbf{E} \in \mathbb{R}^{D\times (n_s + 2r)},\, \mathbf{b}_\text{enc}\in \mathbb{R}^{D},$ and $\sigma$ is a composition of BatchTopK and ReLU \citep{bussmann2024batchtopk}. 
We decode with shared features into both spaces:
\begin{equation}
     \hat{\mathbf{s}}_i = \mathbf{D}^s \, \mathbf{h}_i +\mathbf{b}^s_\text{dec},\quad 
     \hat{\mathbf{z}}_{i,j} = \mathbf{D}^{z_\text{row}} \mathbf{h}_i+ \mathbf{D}^{z_\text{col}}\mathbf{h}_j + \mathbf{b}^z_\text{dec}, \quad i,j\in\{1,\dots, N_\text{tok}\},
\end{equation}
where $\mathbf{D}^s \in \mathbb{R}^{n_s \times D}\,, \mathbf{b}^s_\text{dec}\in \mathbb{R}^{n_s},\,  \mathbf{D}^{z_\text{row}},\mathbf{D}^{z_\text{col}}\in\mathbb{R}^{n_z\times D},$ and $\mathbf{b}^z_\text{dec} \in \mathbb{R}^{n_z}$.

\paragraph{Loss function.} Choosing the dictionary size $D$ can have major consequences on the types of features that are learned, and a larger dictionary does not necessarily translate to better performance on downstream tasks \citep{templeton2024scaling, gaoscaling}. 
It has recently been shown that a Matryoshka SAE loss \citep{bussmannlearning} exhibits robust performance across dictionary sizes. %
This is attained by slicing the decoder matrix and the feature vector across several nested levels (we use three nested widths $c_1<c_2<c_3=D$), and summing them to compute the objectives
\begin{gather}
    \mathcal{L}(\mathbf{s}_i) \coloneq \sum_{k=1}^3\left\| \mathbf{s}_i - \mathbf{D}^s_{[:,\,1:c_k]} \mathbf{h}_{i,[1:c_k]} -\mathbf{b}^s_{\text{dec}} \right\|^2_2 ,\\ \mathcal{L}(\mathbf{z}_{ij}) \coloneq \sum_{k=1}^3\left\| \mathbf{z}_{ij} - \mathbf{D}^{z_\text{row}}_{[:,\,1:c_k]} \mathbf{h}_{i,[1:c_k]}- \mathbf{D}^{z_\text{col}}_{[:,\,1:c_k]} \mathbf{h}_{j,[1:c_k]} -\mathbf{b}^z_{\text{dec}} \right\|^2_2
\end{gather}
Following \citep{gaoscaling}, we include an auxiliary loss function $\mathcal{L}_\text{aux}$ that can "revitalize" dead features, resulting in the loss 
\begin{equation}
    \mathcal{L}(\mathbf{S},\mathcal{Z}) =\sum_{i=1}^{N_\text{tok}}\mathcal{L}(\mathbf{s}_i) + \lambda_\text{pair}\sum_{j=1}^{N_\text{tok}} \mathcal{L}(\mathbf{z}_{ij}) + \lambda_\text{aux} \mathcal{L}_\text{aux}. \
\end{equation}
We speed up training by we computing a Monte Carlo approximation of the second summation and only consider a single $j$ for each $i$. We set $\lambda_\text{pair}= N_\text{tok}^{-1}$ to have both representation be equally weighted.

\begin{figure}
    \centering
    \includegraphics[width=0.49\linewidth]{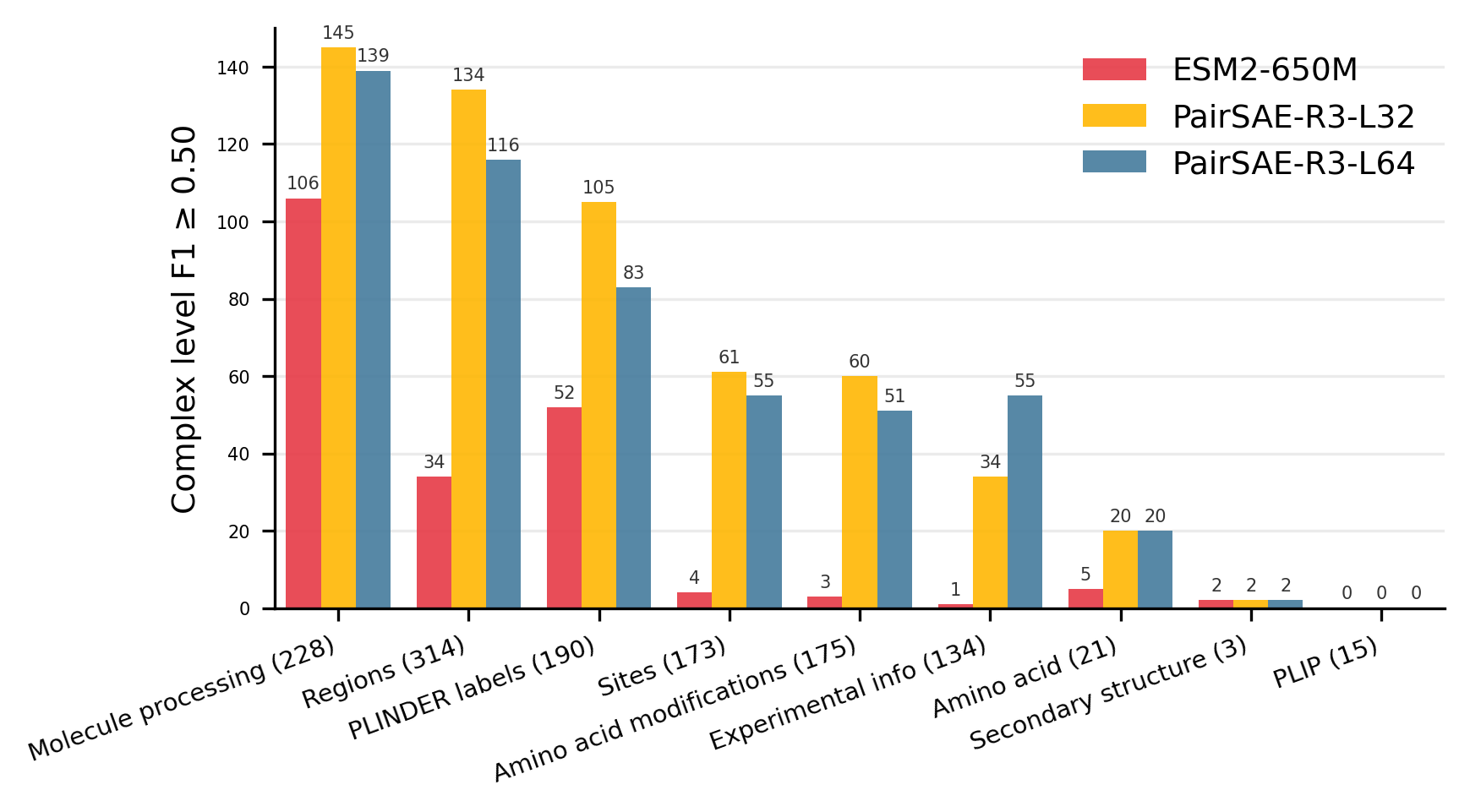}
    \includegraphics[width=0.49\linewidth]{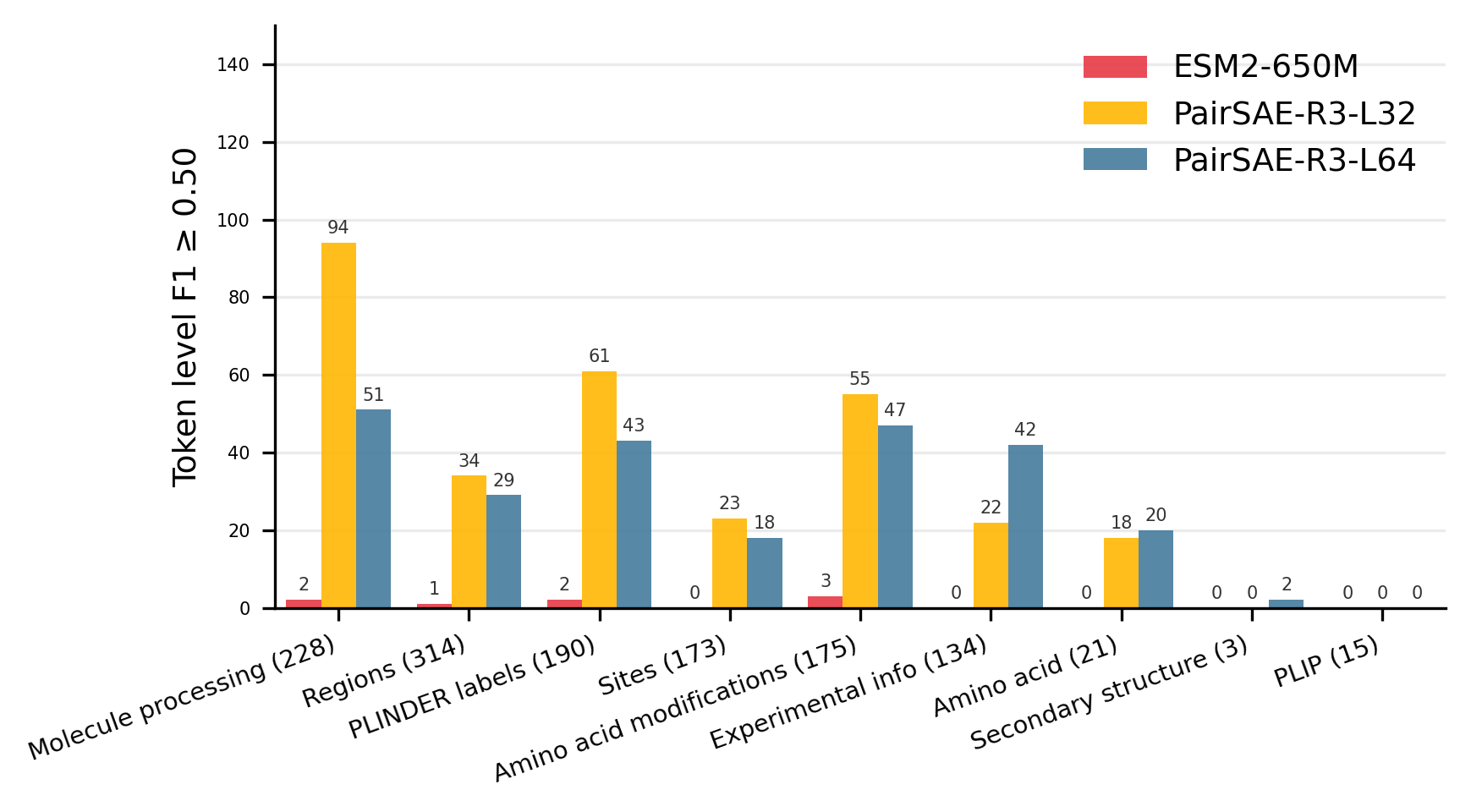}
    \caption{
    Count of concepts with $F_1\ge0.5$ using complex-level recall (left) and token level (right), grouped by UniProt annotation category (test-set counts in parentheses).}
    \label{fig:probes}
\end{figure}
\begin{wraptable}{r}{0.38\linewidth} 
\vspace{-0.5\baselineskip}          
\centering
\footnotesize                           
\setlength{\tabcolsep}{1pt}             
\renewcommand{\arraystretch}{1.05}      
\begin{tabular}{@{}lcc@{}}              
\toprule
Method & Token $F_1$ & Complex $F_1$ \\
\midrule
ESM2-650M        & 0.8\%  & 19.7\% \\
PairSAE-R3-L32   & 29.1\% & 53.2\% \\
PairSAE-R3-L64   & 24.0\% & 49.6\% \\
\bottomrule
\end{tabular}
\caption{Percentage of concepts in the test set with $F_1$ $\geq$ 0.5.}
\label{tab:concept_capture_percentage}
\vspace{-1pt}
\end{wraptable}

\section{Results}
We evaluate PairSAE on Boltz-2 representations for protein–ligand complexes from PLINDER \citep{durairaj2024plinder}. We train two models at the third recycling step, at layers 33 and 64 (R3-L33, R3-L64). Interpretability can be assessed via linear probes \citep{gaoscaling, simon2024interplm}, and we test whether features predict UniProt residue annotations \citep{uniprot2025uniprot} and PLINDER system annotations.

Boltz-2 demonstrated a strong improvement on the speed/accuracy Pareto frontier for binding affinity prediction \citep{passaro2025boltz}, and has the potential of further accelerating virtual screenings. We aim to identify features that exhibit strong association with the predicted affinity, and to do so we build on the hypothesis generation techniques laid out in \cite{movvasparse}. Details of each experiment can be found in Appendix \ref{appdx:details}.

\subsection{Linear probing}
Following \citep{simon2024interplm}, we normalize each feature to $[0,1]$, fit a single-threshold classifier per concept–feature pair via a grid search with candidate thresholds spaced at increments of 0.1 across the unit interval, and select the best feature by validation $F_1$.
As noted in \cite{simon2024interplm}, it is often the case that latent features at a single token might capture properties of the entire system, hence we follow \cite{simon2024interplm} and compute the $F_1$ score by considering the recall at the complex level as well as the token level, resulting in two separate metrics. For each metric, we display a count of how many features we could predict with a score above 0.5 on a test set in Figure \ref{fig:probes}. We find that PairSAE features are highly predictive of UniProt annotations, as shown in Table \ref{tab:concept_capture_percentage}, and outperform classifiers based on neurons from the last layer of ESM-2-650M, considered powerful embeddings for many downstream applications \citep{lin2023evolutionary}.

\subsection{Hypothesis generation: explaining binding affinity predictions}\label{sec:hypo}
Because affinity is defined at the complex level, we form a complex embedding by max-pooling features over tokens: $\hat h_k=\max_{i\le N_\text{tok}} h_{i,k}$, $k=1,\dots,D$. We then predict $AV$ via LASSO \citep{tibshirani1996regression, movvasparse}, minimizing on a training set $\mathcal{D}$ of 980 PLINDER complexes:
\begin{equation}\label{eq:lasso}
    \sum_{j\in\mathcal{D}} \left\| AV_j - \boldsymbol{\beta}\, \mathbf{\hat h}_j \right\|^2_2 + \lambda \left| \boldsymbol{\beta} \right|_1.
\end{equation}
We test our results on Posebusters \citep{buttenschoen2024posebusters}. 

\begin{figure}
    \centering
    \includegraphics[width=0.28\linewidth]{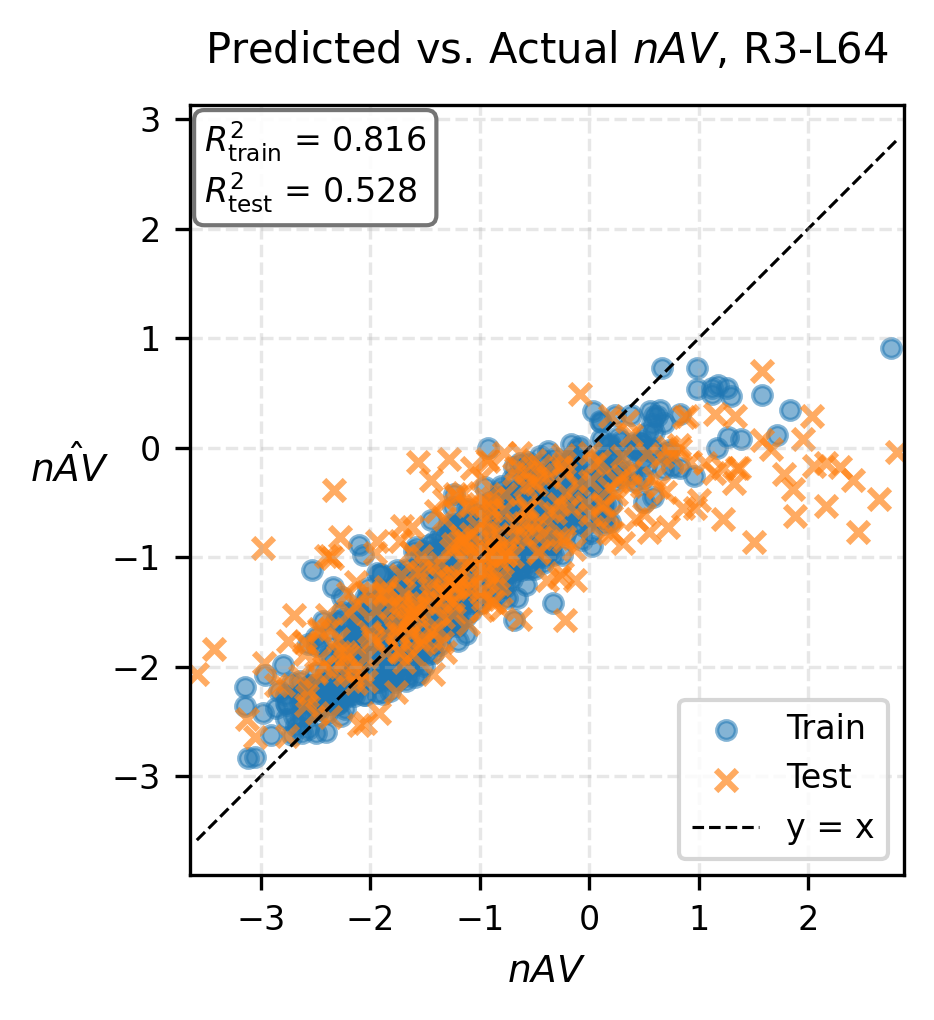}
    \includegraphics[width=0.44\linewidth]{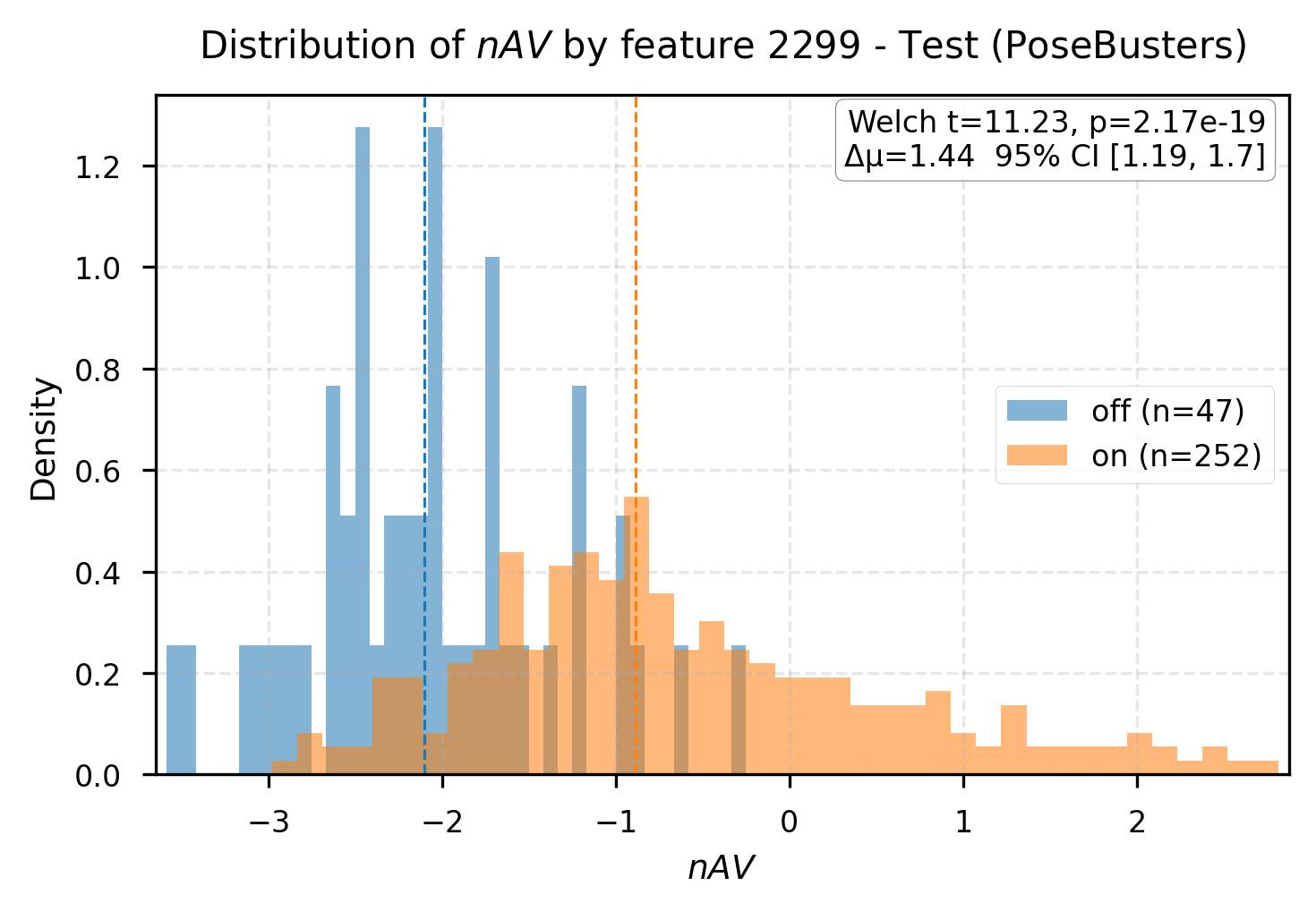}
    \raisebox{1.8em}[0pt][0pt]{%
    \includegraphics[width=0.26\linewidth]{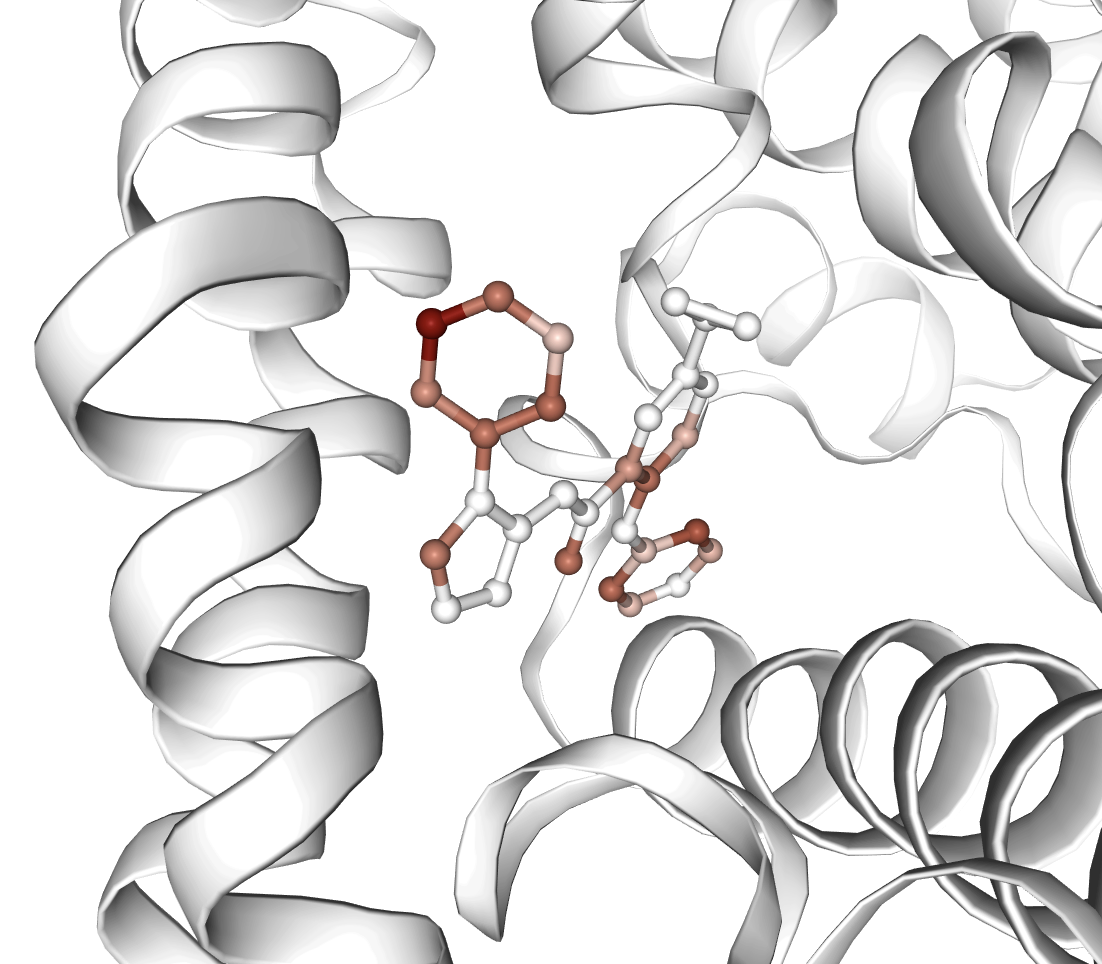}
    }
    \caption{\textbf{Left: } negative affinity values from Boltz-2 (higher means more affine) \citep{passaro2025boltz}, and predicted values from the LASSO regression in \eqref{eq:lasso}. \textbf{Center:} feature 2299 from R3-L64, displaying strong group difference in affinity values measured by Welch t-test. \textbf{Right:} values of feature 2299 overlaid on a ligand where it is highly activated.
    }
    \label{fig:affinity}
\end{figure}

Our first finding is on the predictive power of the PairSAE features on affinity values: by choosing $\lambda$ using cross-validation, we can predict $AV$ with an $R^2$ of 0.528 from R3-L64 using 291 features out of 16,384. 
Errors increase at higher predicted affinity, partly due to label shift (fewer high-affinity examples in training than in PoseBusters; see Fig.~\ref{fig:affinity}).

Moreover, by letting $\lambda$ increase we can highlight features that carry most of the predictive power. By looking at the top 10 most influential features at both layers we find most of them to be activated on ligands, for which we have very limited annotations. In Figure \ref{fig:affinity} we highlight a feature from R3-L64, selected by setting $\lambda$ in \eqref{eq:lasso} such that only one feature has nonzero coefficient. This feature activates on complexes with higher binding affinity, with a strong group difference when compared to inactive complexes. Additional results are reported in Appendix \ref{appdx:additional}.

\section{Conclusion}
We introduced PairSAE, a sparse dictionary learning method that discovers a shared feature basis jointly explaining sequence and pair representations in pairformer-style models. We showed that these features encode biophysically meaningful concepts and can surface hypotheses about a model’s inner workings for downstream tasks such as binding-affinity prediction. Future work includes scaling up the study to more layers and recycling steps, integrating PairSAE features into automated interpretability pipelines to accelerate concept discovery, and developing steering methods for interpretable protein design via feature interventions. A key limitation is that we did not map ligand-activating sparse features to specific concepts, due to the lack of ligand annotations.
\newpage
\bibliographystyle{unsrtnat}
\bibliography{refs}

@inproceedings{movvasparse,
  title={Sparse Autoencoders for Hypothesis Generation},
  author={Movva, Rajiv and Peng, Kenny and Garg, Nikhil and Kleinberg, Jon and Pierson, Emma},
  year={2025},
  booktitle={Forty-second International Conference on Machine Learning}
}

@article{garcia2025interpreting,
  title={Interpreting and Steering Protein Language Models through Sparse Autoencoders},
  author={Garcia, Edith Natalia Villegas and Ansuini, Alessio},
  journal={arXiv preprint arXiv:2502.09135},
  year={2025}
}

@article{adamsmechanistic,
  title={From Mechanistic Interpretability to Mechanistic Biology: Training, Evaluating, and Interpreting Sparse Autoencoders on Protein Language Models},
  author={Adams, Etowah and Bai, Liam and Lee, Minji and Yu, Yiyang and AlQuraishi, Mohammed},
  year={2025},
  journal={Forty-second International Conference on Machine Learning}
}

@article{gujral2025sparse,
  title={Sparse autoencoders uncover biologically interpretable features in protein language model representations},
  author={Gujral, Onkar and Bafna, Mihir and Alm, Eric and Berger, Bonnie},
  journal={Proceedings of the National Academy of Sciences},
  volume={122},
  number={34},
  pages={e2506316122},
  year={2025},
  publisher={National Academy of Sciences}
}

@article{uniprot2025uniprot,
  title={UniProt: the universal protein knowledgebase in 2025},
  author={UniProt},
  journal={Nucleic acids research},
  volume={53},
  number={D1},
  pages={D609--D617},
  year={2025},
  publisher={Oxford University Press}
}

@article{kingma2014adam,
  title={Adam: A method for stochastic optimization},
  author={Kingma, Diederik P and Ba, Jimmy},
  journal={arXiv preprint arXiv:1412.6980},
  year={2014}
}

@article{scikit-learn,
  title={Scikit-learn: Machine Learning in {P}ython},
  author={Pedregosa, F. and Varoquaux, G. and Gramfort, A. and Michel, V.
          and Thirion, B. and Grisel, O. and Blondel, M. and Prettenhofer, P.
          and Weiss, R. and Dubourg, V. and Vanderplas, J. and Passos, A. and
          Cournapeau, D. and Brucher, M. and Perrot, M. and Duchesnay, E.},
  journal={Journal of Machine Learning Research},
  volume={12},
  pages={2825--2830},
  year={2011}
}

@article{velankar2012sifts,
  title={SIFTS: structure integration with function, taxonomy and sequences resource},
  author={Velankar, Sameer and Dana, Jos{\'e} M and Jacobsen, Julius and Van Ginkel, Glen and Gane, Paul J and Luo, Jie and Oldfield, Thomas J and O’Donovan, Claire and Martin, Maria-Jesus and Kleywegt, Gerard J},
  journal={Nucleic acids research},
  volume={41},
  number={D1},
  pages={D483--D489},
  year={2012},
  publisher={Oxford University Press}
}

@article{salentin2015plip,
  title={PLIP: fully automated protein--ligand interaction profiler},
  author={Salentin, Sebastian and Schreiber, Sven and Haupt, V Joachim and Adasme, Melissa F and Schroeder, Michael},
  journal={Nucleic acids research},
  volume={43},
  number={W1},
  pages={W443--W447},
  year={2015},
  publisher={Oxford University Press}
}

@article{buttenschoen2024posebusters,
  title={PoseBusters: AI-based docking methods fail to generate physically valid poses or generalise to novel sequences},
  author={Buttenschoen, Martin and Morris, Garrett M and Deane, Charlotte M},
  journal={Chemical Science},
  volume={15},
  number={9},
  pages={3130--3139},
  year={2024},
  publisher={Royal Society of Chemistry}
}

@article{lin2023evolutionary,
  title={Evolutionary-scale prediction of atomic-level protein structure with a language model},
  author={Lin, Zeming and Akin, Halil and Rao, Roshan and Hie, Brian and Zhu, Zhongkai and Lu, Wenting and Smetanin, Nikita and Verkuil, Robert and Kabeli, Ori and Shmueli, Yaniv and others},
  journal={Science},
  volume={379},
  number={6637},
  pages={1123--1130},
  year={2023},
  publisher={American Association for the Advancement of Science}
}

@article{tibshirani1996regression,
  title={Regression shrinkage and selection via the lasso},
  author={Tibshirani, Robert},
  journal={Journal of the Royal Statistical Society Series B: Statistical Methodology},
  volume={58},
  number={1},
  pages={267--288},
  year={1996},
  publisher={Oxford University Press}
}

@article{lee2006efficient,
  title={Efficient sparse coding algorithms},
  author={Lee, Honglak and Battle, Alexis and Raina, Rajat and Ng, Andrew},
  journal={Advances in neural information processing systems},
  volume={19},
  year={2006}
}

@inproceedings{parsan2025towards,
  title={Towards Interpretable Protein Structure Prediction with Sparse Autoencoders},
  author={Parsan, Nithin and Yang, David J and Yang, John Jingxuan},
  booktitle={Learning Meaningful Representations of Life (LMRL) Workshop at ICLR 2025},
  year={2025}
}

@article{jumper2021highly,
  title={Highly accurate protein structure prediction with AlphaFold},
  author={Jumper, John and Evans, Richard and Pritzel, Alexander and Green, Tim and Figurnov, Michael and Ronneberger, Olaf and Tunyasuvunakool, Kathryn and Bates, Russ and {\v{Z}}{\'\i}dek, Augustin and Potapenko, Anna and others},
  journal={nature},
  volume={596},
  number={7873},
  pages={583--589},
  year={2021},
  publisher={Nature Publishing Group UK London}
}

@article{abramson2024accurate,
  title={Accurate structure prediction of biomolecular interactions with AlphaFold 3},
  author={Abramson, Josh and Adler, Jonas and Dunger, Jack and Evans, Richard and Green, Tim and Pritzel, Alexander and Ronneberger, Olaf and Willmore, Lindsay and Ballard, Andrew J and Bambrick, Joshua and others},
  journal={Nature},
  volume={630},
  number={8016},
  pages={493--500},
  year={2024},
  publisher={Nature Publishing Group UK London}
}

@article{wohlwend2025boltz,
  title={Boltz-1 democratizing biomolecular interaction modeling},
  author={Wohlwend, Jeremy and Corso, Gabriele and Passaro, Saro and Getz, Noah and Reveiz, Mateo and Leidal, Ken and Swiderski, Wojtek and Atkinson, Liam and Portnoi, Tally and Chinn, Itamar and others},
  journal={BioRxiv},
  pages={2024--11},
  year={2025}
}

@article{passaro2025boltz,
  title={Boltz-2: Towards accurate and efficient binding affinity prediction},
  author={Passaro, Saro and Corso, Gabriele and Wohlwend, Jeremy and Reveiz, Mateo and Thaler, Stephan and Somnath, Vignesh Ram and Getz, Noah and Portnoi, Tally and Roy, Julien and Stark, Hannes and others},
  journal={BioRxiv},
  pages={2025--06},
  year={2025},
  publisher={Cold Spring Harbor Laboratory}
}

@article{ng2011sparse,
  title={Sparse autoencoder},
  author={Ng, Andrew and others},
  journal={CS294A Lecture notes},
  volume={72},
  number={2011},
  pages={1--19},
  year={2011}
}

@article{olshausen1997sparse,
  title={Sparse coding with an overcomplete basis set: A strategy employed by V1?},
  author={Olshausen, Bruno A and Field, David J},
  journal={Vision research},
  volume={37},
  number={23},
  pages={3311--3325},
  year={1997},
  publisher={Elsevier}
}

@article{durairaj2024plinder,
  title={PLINDER: The protein-ligand interactions dataset and evaluation resource},
  author={Durairaj, Janani and Adeshina, Yusuf and Cao, Zhonglin and Zhang, Xuejin and Oleinikovas, Vladas and Duignan, Thomas and McClure, Zachary and Robin, Xavier and Studer, Gabriel and Kovtun, Daniel and others},
  journal={bioRxiv},
  pages={2024--07},
  year={2024},
  publisher={Cold Spring Harbor Laboratory}
}

@inproceedings{gaoscaling,
  title={Scaling and evaluating sparse autoencoders},
  author={Gao, Leo and la Tour, Tom Dupre and Tillman, Henk and Goh, Gabriel and Troll, Rajan and Radford, Alec and Sutskever, Ilya and Leike, Jan and Wu, Jeffrey},
  year={2025},
  booktitle={The Thirteenth International Conference on Learning Representations}
}

@misc{bricken2023monosemanticity,
  title        = {Towards monosemanticity: Decomposing language models with dictionary learning},
  author       = {Trenton Bricken and Adly Templeton and Joshua Batson and Brian Chen and Adam Jermyn and Tom Conerly and Nick Turner and Cem Anil and Carson Denison and Amanda Askell and Robert Lasenby and Yifan Wu and Shauna Kravec and Nicholas Schiefer and Tim Maxwell and Nicholas Joseph and Zac Hatfield-Dodds and Alex Tamkin and Karina Nguyen and Brayden McLean and Josiah E. Burke and Tristan Hume and Shan Carter and Tom Henighan and Christopher Olah},
  year         = {2023},
  howpublished = {\url{https://transformer-circuits.pub/2023/monosemantic-features/index.html}},
  note         = {Transformer Circuits Thread}
}

@article{elhage2022toy,
  title={Toy models of superposition},
  author={Elhage, Nelson and Hume, Tristan and Olsson, Catherine and Schiefer, Nicholas and Henighan, Tom and Kravec, Shauna and Hatfield-Dodds, Zac and Lasenby, Robert and Drain, Dawn and Chen, Carol and others},
  journal={arXiv preprint arXiv:2209.10652},
  year={2022}
}

@article{cunningham2023sparseautoencodershighlyinterpretable,
      title={Sparse Autoencoders Find Highly Interpretable Features in Language Models}, 
      author={Hoagy Cunningham and Aidan Ewart and Logan Riggs and Robert Huben and Lee Sharkey},
      year={2024},
      journal={ICLR}
}

@article{scherlis2022polysemanticity,
  title={Polysemanticity and capacity in neural networks},
  author={Scherlis, Adam and Sachan, Kshitij and Jermyn, Adam S and Benton, Joe and Shlegeris, Buck},
  journal={arXiv preprint arXiv:2210.01892},
  year={2022}
}

@inproceedings{vasilescu2002multilinear,
  title={Multilinear analysis of image ensembles: Tensorfaces},
  author={Vasilescu, M Alex O and Terzopoulos, Demetri},
  booktitle={European conference on computer vision},
  pages={447--460},
  year={2002},
  organization={Springer}
}

@article{de2000multilinear,
  title={A multilinear singular value decomposition},
  author={De Lathauwer, Lieven and De Moor, Bart and Vandewalle, Joos},
  journal={SIAM journal on Matrix Analysis and Applications},
  volume={21},
  number={4},
  pages={1253--1278},
  year={2000},
  publisher={SIAM}
}

@article{de2000best,
  title={On the best rank-1 and rank-($R_1, R_2,..., R_n$) approximation of higher-order tensors},
  author={De Lathauwer, Lieven and De Moor, Bart and Vandewalle, Joos},
  journal={SIAM journal on Matrix Analysis and Applications},
  volume={21},
  number={4},
  pages={1324--1342},
  year={2000},
  publisher={SIAM}
}

@article{templeton2024scaling,
  title        = {Scaling Monosemanticity: Extracting Interpretable Features from Claude 3 Sonnet},
  author       = {Templeton, Adly and Conerly, Tom and Marcus, Jonathan and Lindsey, Jack and 
                  Bricken, Trenton and Chen, Brian and Pearce, Adam and Citro, Craig and Ameisen, Emmanuel and
                  Jones, Andy and Cunningham, Hoagy and Turner, Nicholas L. and McDougall, Callum and
                  MacDiarmid, Monte and Freeman, C. Daniel and Sumers, Theodore R. and Rees, Edward and
                  Batson, Joshua and Jermyn, Adam and Carter, Shan and Olah, Chris and Henighan, Tom},
  year         = {2024},
  journal      = {Transformer Circuits Thread},
  url          = {https://transformer-circuits.pub/2024/scaling-monosemanticity/index.html}
}

@article{simon2024interplm,
  title={Interplm: Discovering interpretable features in protein language models via sparse autoencoders},
  author={Simon, Elana and Zou, James},
  journal={bioRxiv},
  pages={2024--11},
  year={2024},
  publisher={Cold Spring Harbor Laboratory}
}

@article{bussmann2024batchtopk,
  title={Batchtopk sparse autoencoders},
  author={Bussmann, Bart and Leask, Patrick and Nanda, Neel},
  journal={arXiv preprint arXiv:2412.06410},
  year={2024}
}

@inproceedings{bussmannlearning,
  title={Learning Multi-Level Features with Matryoshka Sparse Autoencoders},
  author={Bussmann, Bart and Nabeshima, Noa and Karvonen, Adam and Nanda, Neel},
  booktitle={Forty-second International Conference on Machine Learning},
  year={2025}
}

@article{ishteva2011tucker,
  title={Tucker compression and local optima},
  author={Ishteva, Mariya and Absil, P-A and Van Huffel, Sabine and De Lathauwer, Lieven},
  journal={Chemometrics and Intelligent Laboratory Systems},
  volume={106},
  number={1},
  pages={57--64},
  year={2011},
  publisher={Elsevier}
}

@article{vannieuwenhoven2012new,
  title={A new truncation strategy for the higher-order singular value decomposition},
  author={Vannieuwenhoven, Nick and Vandebril, Raf and Meerbergen, Karl},
  journal={SIAM Journal on Scientific Computing},
  volume={34},
  number={2},
  pages={A1027--A1052},
  year={2012},
  publisher={SIAM}
}

@article{brixi2025genome,
  title={Genome modeling and design across all domains of life with Evo 2},
  author={Brixi, Garyk and Durrant, Matthew G and Ku, Jerome and Poli, Michael and Brockman, Greg and Chang, Daniel and Gonzalez, Gabriel A and King, Samuel H and Li, David B and Merchant, Aditi T and others},
  journal={BioRxiv},
  pages={2025--02},
  year={2025},
  publisher={Cold Spring Harbor Laboratory}
}

\newpage
\appendix
\renewcommand{\thefigure}{\Alph{section}.\arabic{figure}}
\renewcommand{\thetable}{\Alph{section}.\arabic{table}}
\section{$N$-mode Singular Value Decomposition}\label{appdx:svd} 

The mode-$k$ unfolding of a tensor $\mathcal{T}\in \mathbb{R}^{n_1\times n_2\times \dots \times n_N}$ is obtained by flattening all but the $k^\text{th}$ dimension into a matrix $\mathcal{T}^{(k)}\in\mathbb{R}^{n_k\times \prod_{i\neq k} n_i}$. Following \cite{de2000multilinear}, every tensor admits the higher-order singular value decomposition
\[
    \mathcal{T}=\mathcal{C}\times_1 \mathbf{U}^{(1)}\times_2 \mathbf{U}^{(2)}\dots \times_N \mathbf{U}^{(N)},
\]
where $\mathcal{C}\in \mathbb{R}^{n_1\times n_2\times \dots \times n_N}$ is a core tensor, $\times_k$ denotes the mode-$k$ tensor–matrix product, and each $\mathbf{U}^{(k)} \in \mathbb{R}^{n_k\times n_k}$ is an orthogonal matrix obtained from the left-singular vectors of the mode-$k$ unfolding $\mathcal{T}^{(k)}$ \citep{vasilescu2002multilinear}. We refer to $\mathbf{U}^{(k)}$ as the mode-$k$ left singular vectors of $\mathcal{T}$.

Unlike matrices, truncating the ordered left-singular vectors to their $r_k < n_k$ columns and fitting a new, smaller core tensor does not yield an optimal ($r_1,r_2,\dots,r_N$)-rank approximation of $\mathcal{T}$ in the Frobenius norm \citep{de2000best}. Multiple algorithms have been proposed to improve upon a naive truncation \citep{de2000best, vannieuwenhoven2012new}, but convergence guarantees are limited to local optima \citep{ishteva2011tucker}.

\section{Experimental Details}\label{appdx:details}

\subsection{PairSAE}
Following the standard implementation of Boltz-2, our sequence representation has dimension $n_s=384$, while the pair representation has dimension $n_z=128$. We compute the $N-$mode SVD by simply flattening $\mathcal{Z}$ to its mode-1 and mode-2 unfolding, and obtain the SVD of these matrices using \texttt{numpy.linalg.svd}. We truncate to the first $r=64$ columns, and if $N_\text{tok}<r$ we fill the remaining entries with zeroes. After concatenating sequence embeddings $\mathbf{s}$ and the SVD-derived embedding $\mathbf{m}$ into a 512-dimensional vector, we perform a layer normalization.

We let PairSAE have a dictionary size of $D=16,384$, corresponding to an expansion factor of $32\times$. We train on minibatches of 2,048 tokens using the Adam optimizer \citep{kingma2014adam}, with learning rate 0.0002. We train for 250,000 steps on 40,000 systems sampled from the PLINDER training set \citep{durairaj2024plinder} with at most 512 residues, and at each step we sample tokens at random. Due to compute constraints, we do not use the MSA when training and evaluating the PairSAE.

\subsection{Linear probing}

We consider annotations from UniProt \citep{uniprot2025uniprot}, and binarize them by one-hot encoding each annotation. We map each chain in a PLINDER system to its UniProt annotation using the SIFTS database \citep{velankar2012sifts}, and match sequences with a minimum of 95\% correspondance between the two databases, ignoring sequences that can't be matched.

We also consider a subset of the system-level annotations found in PLINDER, as well as the PLIP interaction fingerprints \citep{salentin2015plip} that are also found in PLINDER. Iterating over features $h_i$ for $i=1,\dots,D$ and concept annotations $y_j$ for $j=1,\dots,N_y$, we consider a simple classifier
\begin{equation}
\hat y_j = \boldsymbol{1}\{h_i >\tau_{ij} \}.
\end{equation}
We pick the best threshold $\tau_{ij}$ on a training set of 7,680 complexes from the PLINDER training set, we select the best feature $i$ for each concept $j$ based on $F_1$ scores on a validation set of 2,560 complexes, and finally report $F_1$ scores on a test set comprised of 5,120 randomly selected complexes. We restrict our analysis to concepts that appear on at least five different complexes.

\subsection{Hypothesis generation}
\begin{figure}
    \centering
    \includegraphics[width=0.5\linewidth]{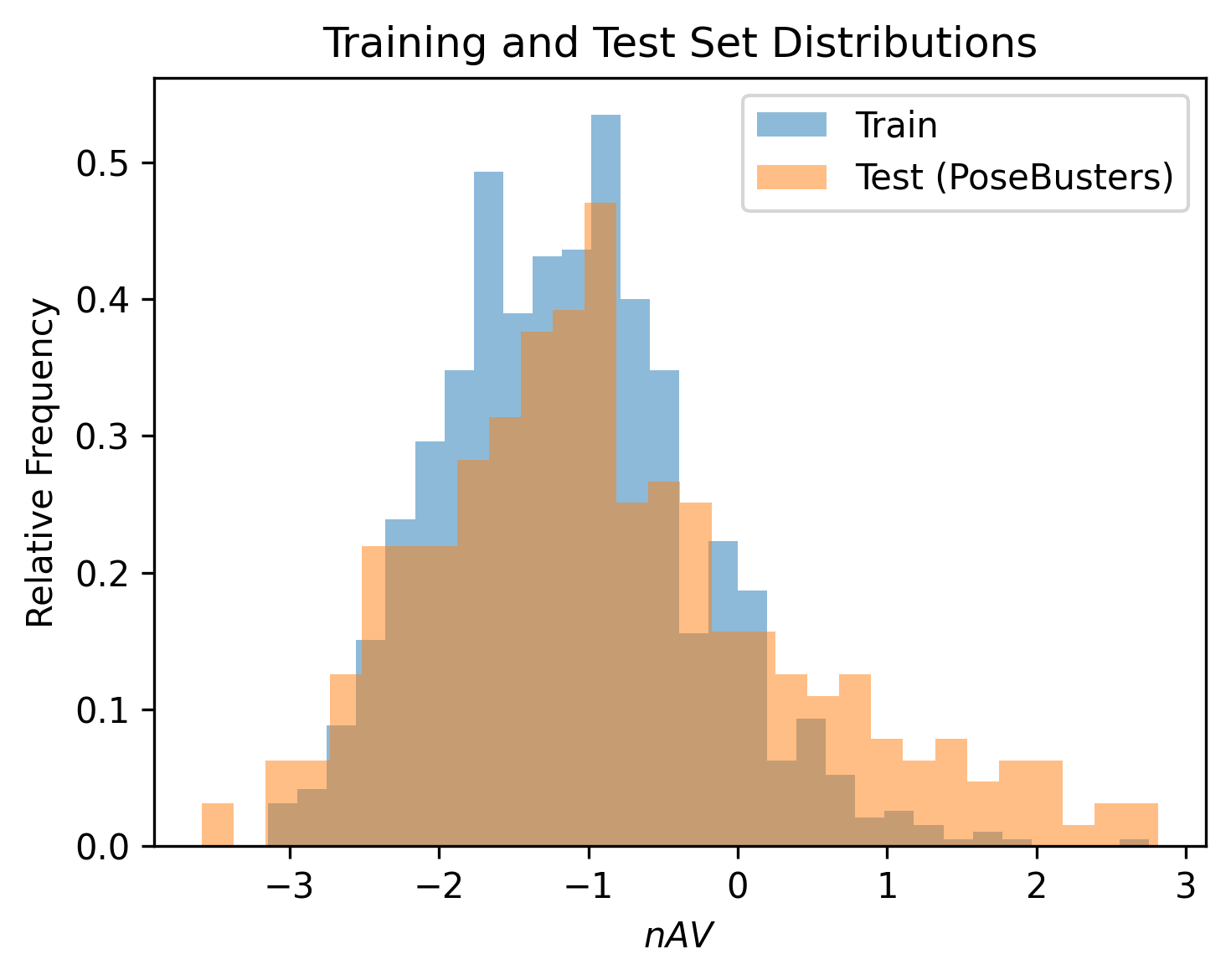}
    \caption{Boltz-2 affinity values for the training and test set used in the hypothesis generation task.}
    \label{fig:train_test}
\end{figure}

We build a training set for LASSO regression by obtaining Boltz-2 predictions for affinity values in 980 systems from PLINDER, held out from the PairSAE training set. The test set is Posebusters \citep{buttenschoen2024posebusters}. In figure \ref{fig:train_test} we highlight how these differ in terms of the response variable, and note the lack of high affinity complexes in our training set.

\begin{figure}
    \centering
    \includegraphics[width=0.4\linewidth]{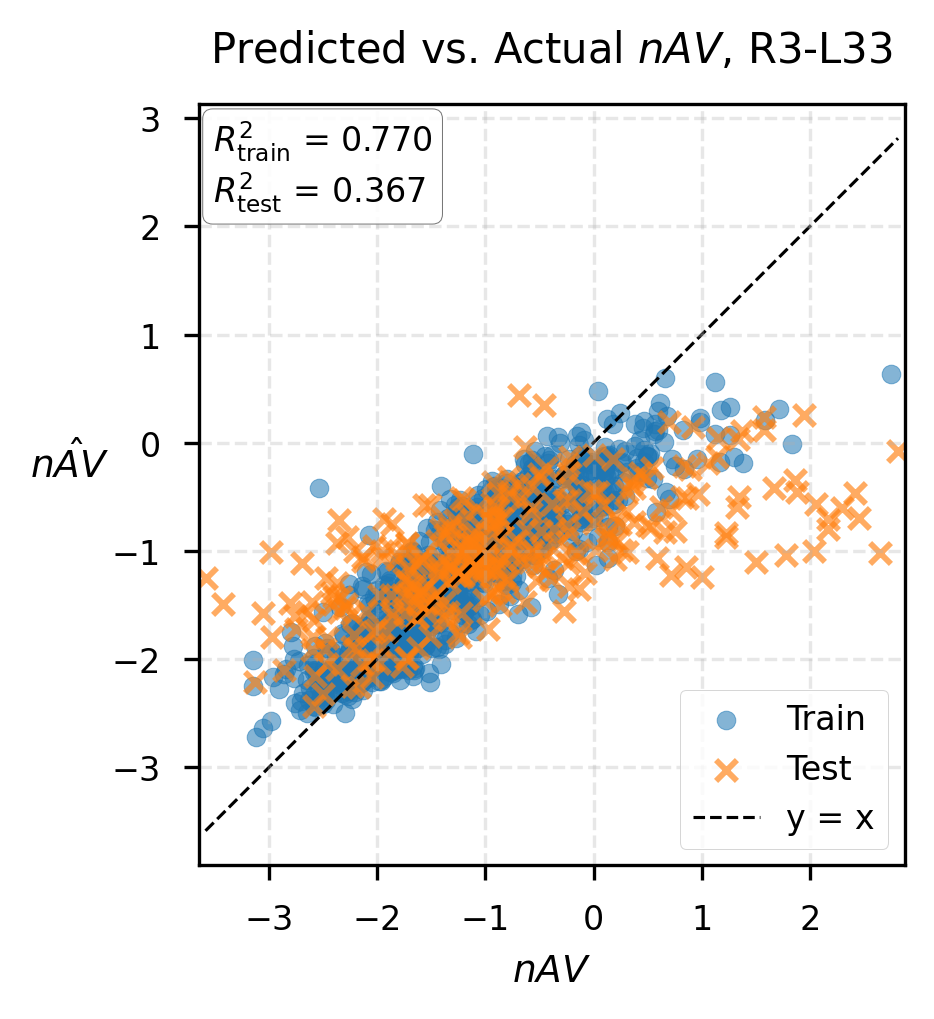}
    \caption{Negative affinity values from Boltz-2 (higher means more affine), and predicted values from the LASSO regression in \eqref{eq:lasso} using the PairSAE at R3-L33.}
    \label{fig:lasso_33}
\end{figure}

LASSO regression is fit using \texttt{sklearn.linear\_model.Lasso} \citep{scikit-learn}. Detailed results for both models are presented in Table \ref{tab:lasso_metrics}. See Figure \ref{fig:lasso_33} for true vs predicted Boltz-2 affinity values based on the PairSAE at R3-L33.

\begin{figure}
    \centering
    \includegraphics[width=0.9\linewidth]{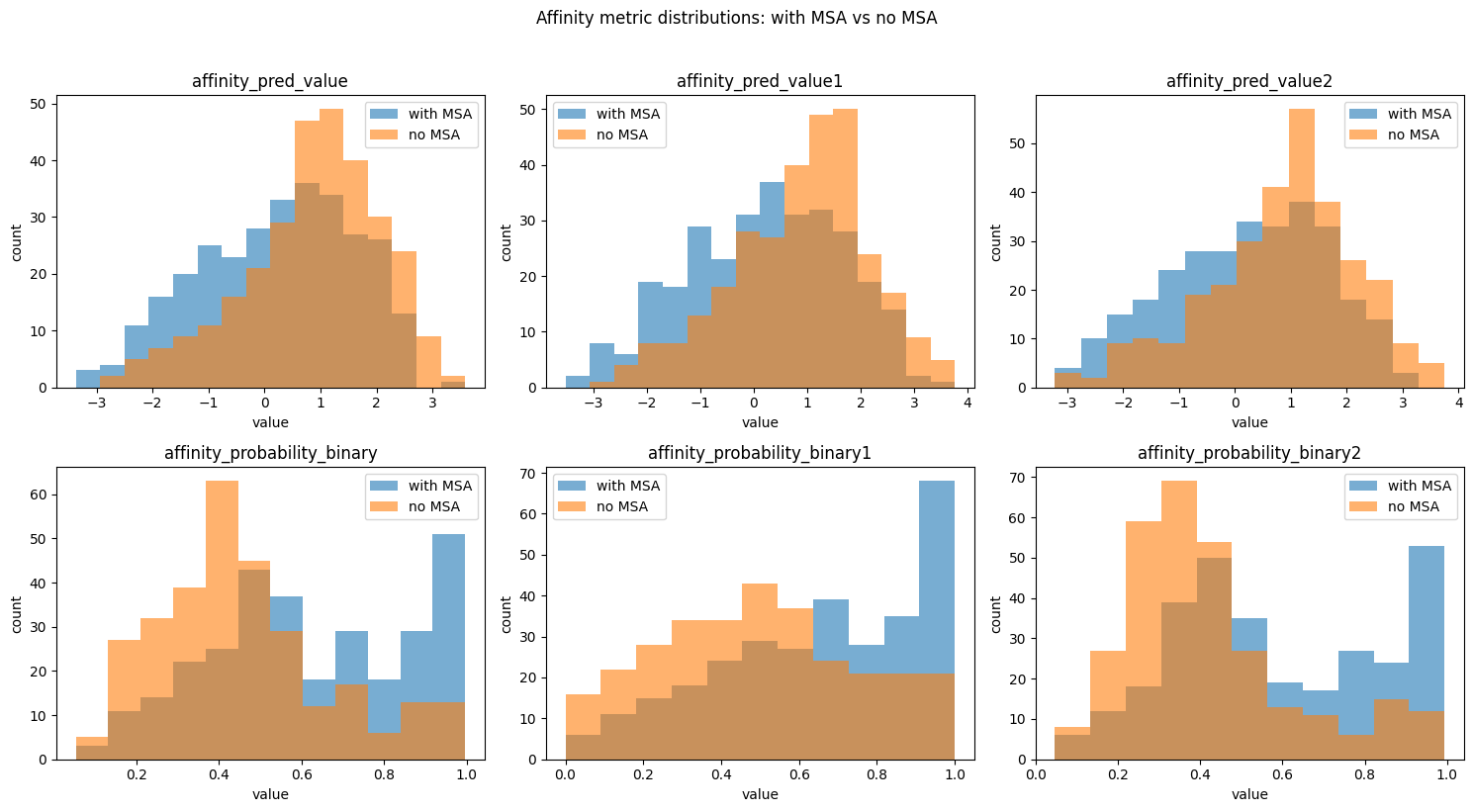}
    \caption{Comparison of having MSA on and off in the Boltz-2 affinity values and affinity probabilities, in the Posebusters dataset.}
    \label{fig:msa}
\end{figure}

As mentioned in Section \ref{appdx:details}, we did not use the MSA for our analysis. In Figure \ref{fig:msa} we compare the Boltz-2 output with and without MSA, and note a substantial difference in predicted affinity values and affinity probability. As MSA-based predictions are conditioned on more information, we expect these to be more accurate. In future work, we aim at replicating our analysis using MSA inputs.

\begin{table}[h]
\centering
\footnotesize
\setlength{\tabcolsep}{4pt}
\renewcommand{\arraystretch}{1.1}
\caption{LASSO regression on max-pooled PairSAE features. $\rho$ denotes Spearman rank correlation.}

\begin{tabular}{@{}lccccc ccc c@{}}
\toprule
& & \multicolumn{5}{c}{\textbf{Prediction metrics}} & \multicolumn{2}{c}{\textbf{Ranking}} & \multicolumn{1}{c}{\textbf{Sparsity}} \\
\cmidrule(lr){3-7} \cmidrule(lr){8-9} \cmidrule(lr){10-10}
\textbf{Model} & $\lambda$ (CV) & Train $R^2$ & Train MAE & Train RMSE & Test $R^2$ & Test MAE & Train $\rho$ & Test $\rho$ & Nonzero \\
\midrule
R3--L64 & 0.009 & 0.816 & 0.275 & 0.131 & 0.528 & 0.607 & 0.913 & 0.805 & 291 \\
R3--L33 & 0.006 & 0.770 & 0.302 & 0.164 & 0.367 & 0.713 & 0.893 & 0.706 & 237 \\
\bottomrule
\end{tabular}
\label{tab:lasso_metrics}
\end{table}


\newpage
\section{Additional results}\label{appdx:additional}
\begin{figure}
    \centering
    \includegraphics[width=0.5\linewidth]{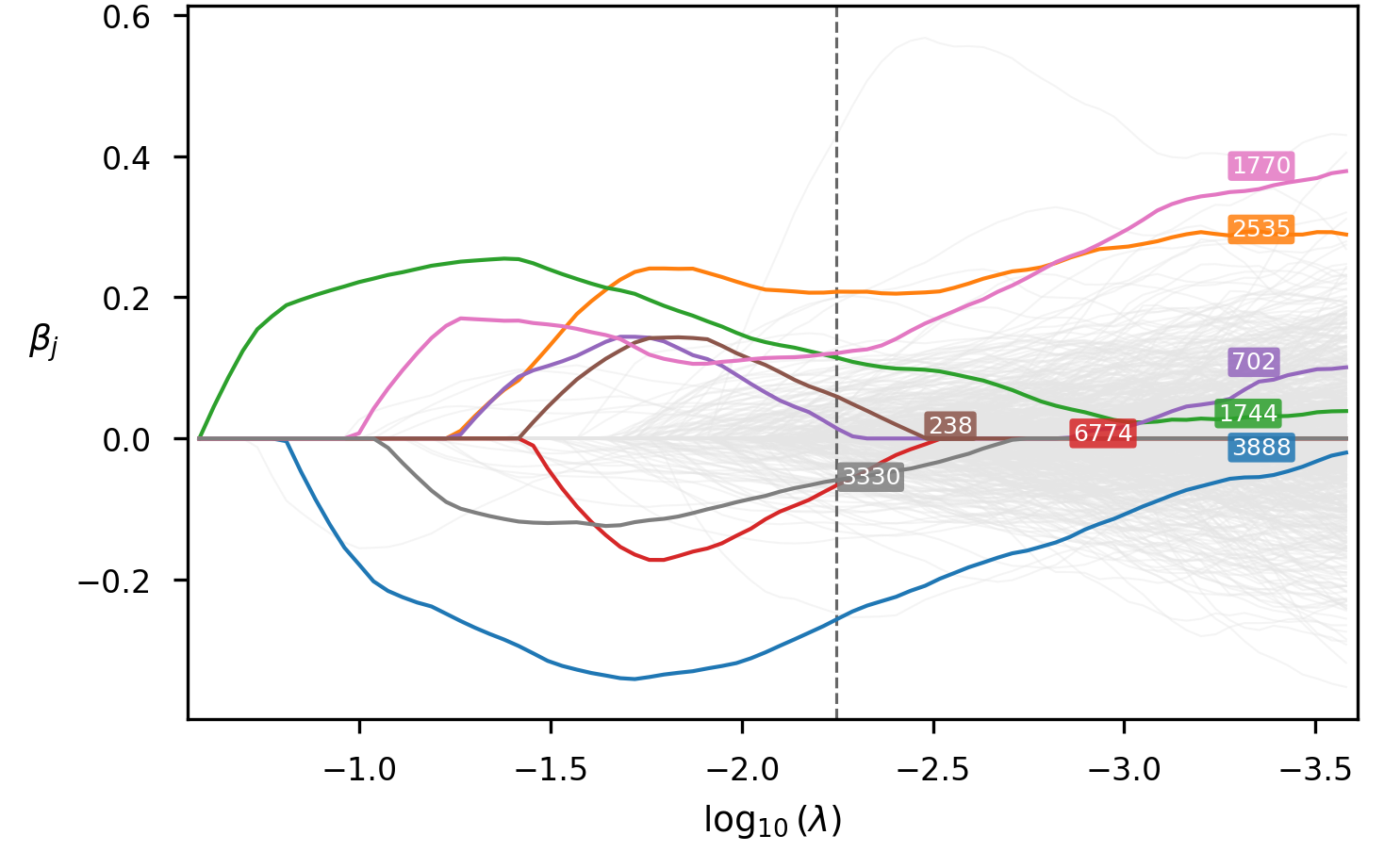}\includegraphics[width=0.5\linewidth]{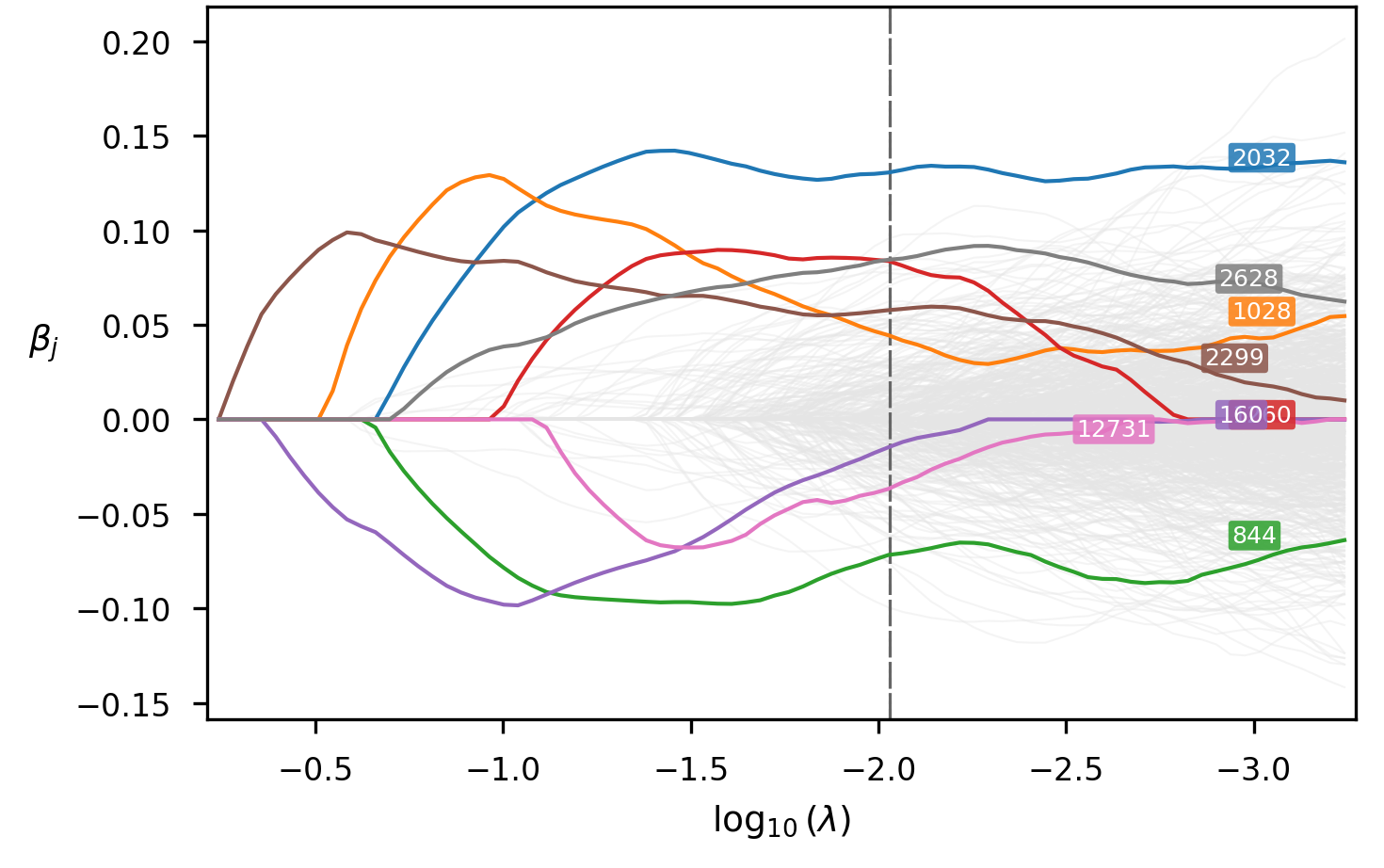}
    \caption{LASSO regression coefficients for varying values of the $\lambda$ regularization parameter in \eqref{eq:lasso}, fit on the PairSAE at R3-L33 (left) and R3-L64 (right). }
    \label{fig:lasso_paths}
\end{figure}

\begin{figure}
    \centering
    \includegraphics[width=0.45\linewidth]{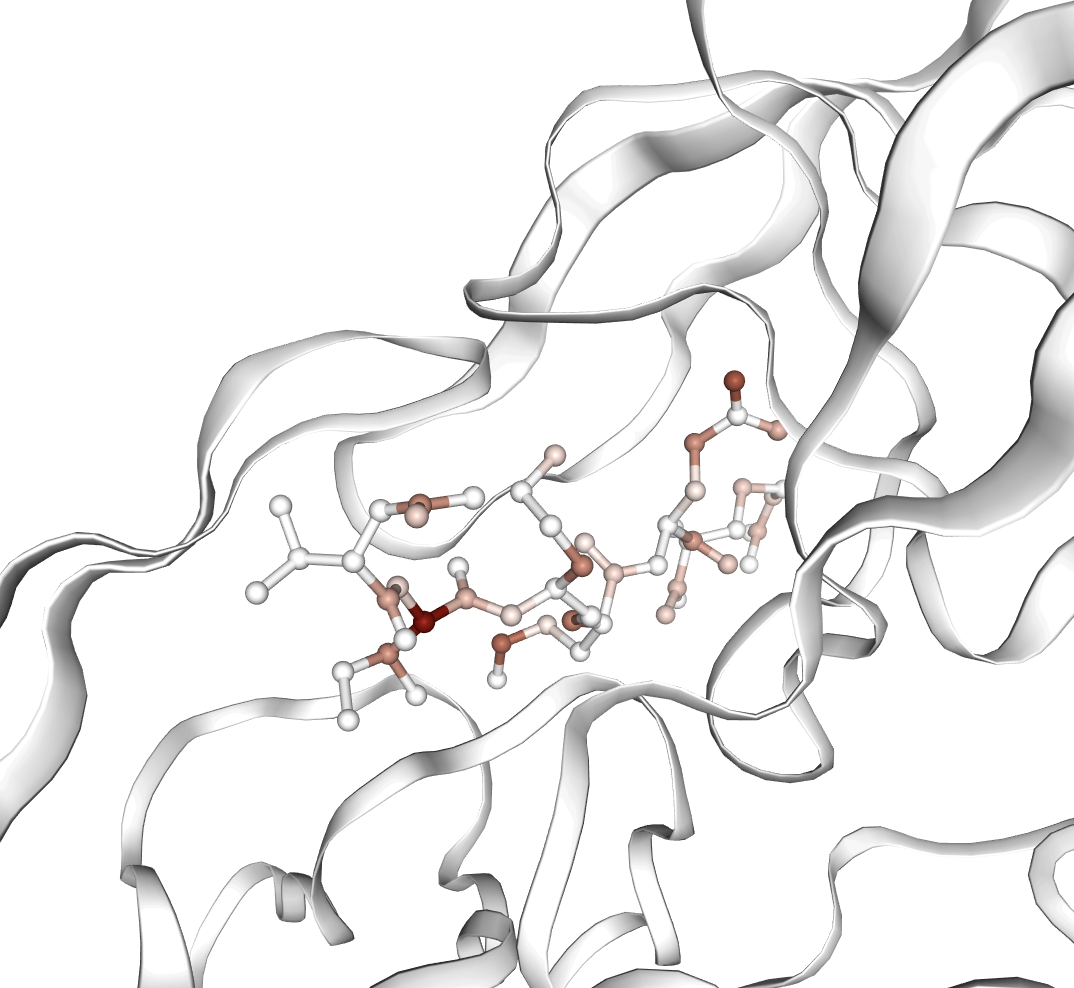}
\hspace{5pt}
    \includegraphics[width=0.45\linewidth]{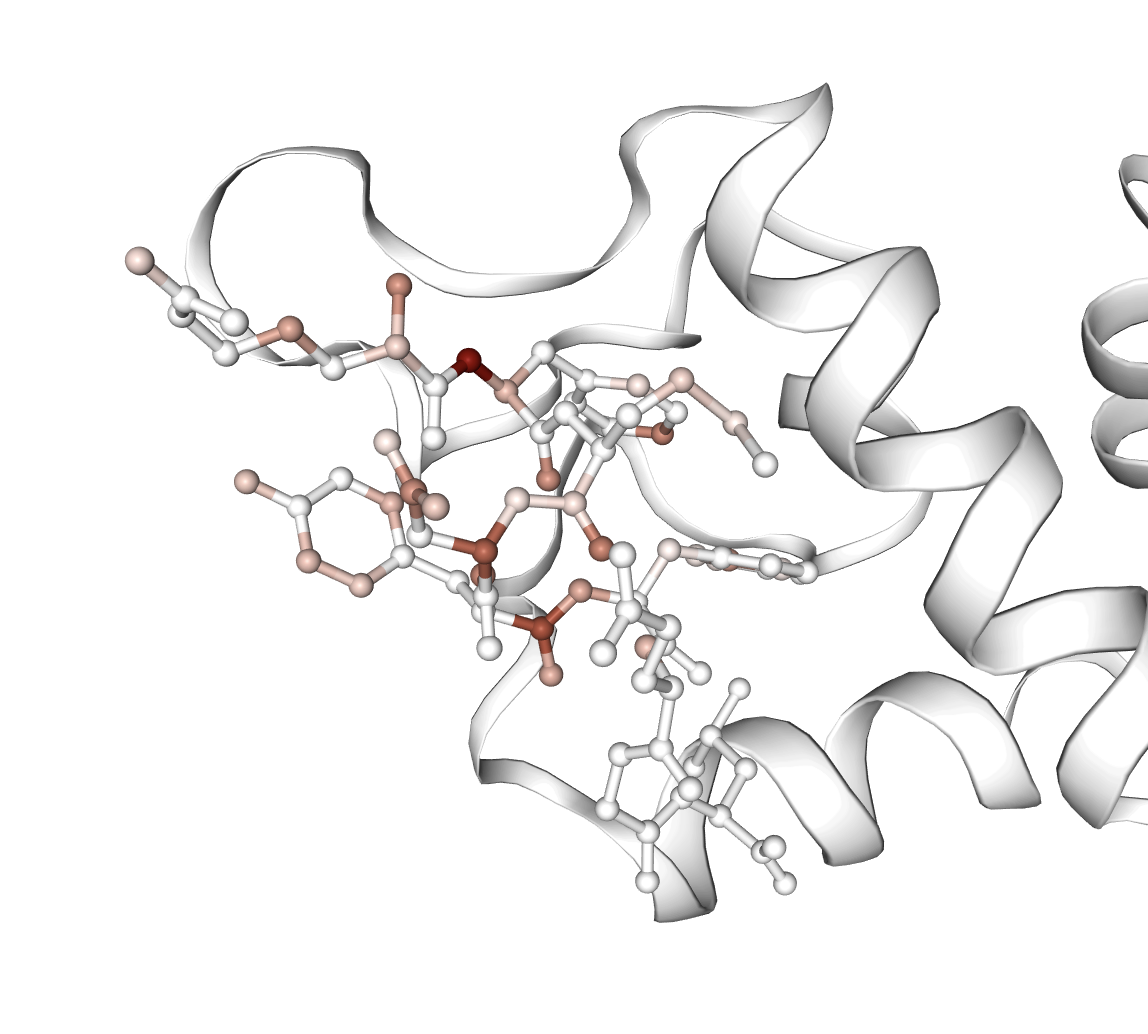}
    \caption{Ligands activating on feature 1770 (left) and feature 1774 (right). These are the top 2 positive features that predict the Boltz-2 affinity in R3-L33, as highlighted in Figure \ref{fig:lasso_paths}. }
    \label{fig:ligands}
\end{figure}

In Figure \ref{fig:lasso_paths} we display how lasso coefficients evolve for varying levels of regularization, and highlight features having strong association with the predicted affinity. Among these, we highlight how feature 1744 and 1770 from R3-L33 activate in two ligands in Figure \ref{fig:ligands}.

\begin{figure}
    \centering
    \includegraphics[width=0.49\linewidth]{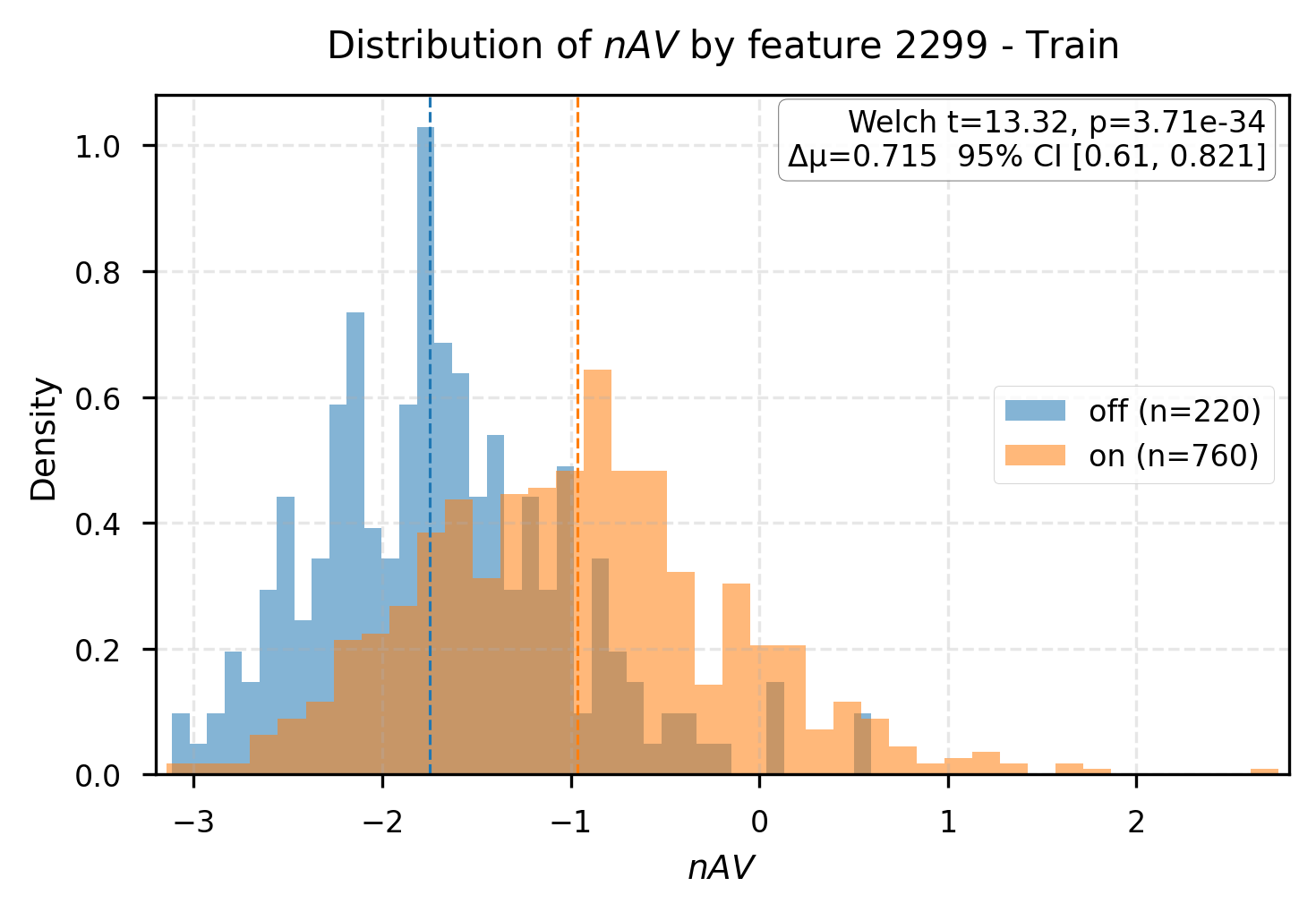}
    \includegraphics[width=0.49\linewidth]{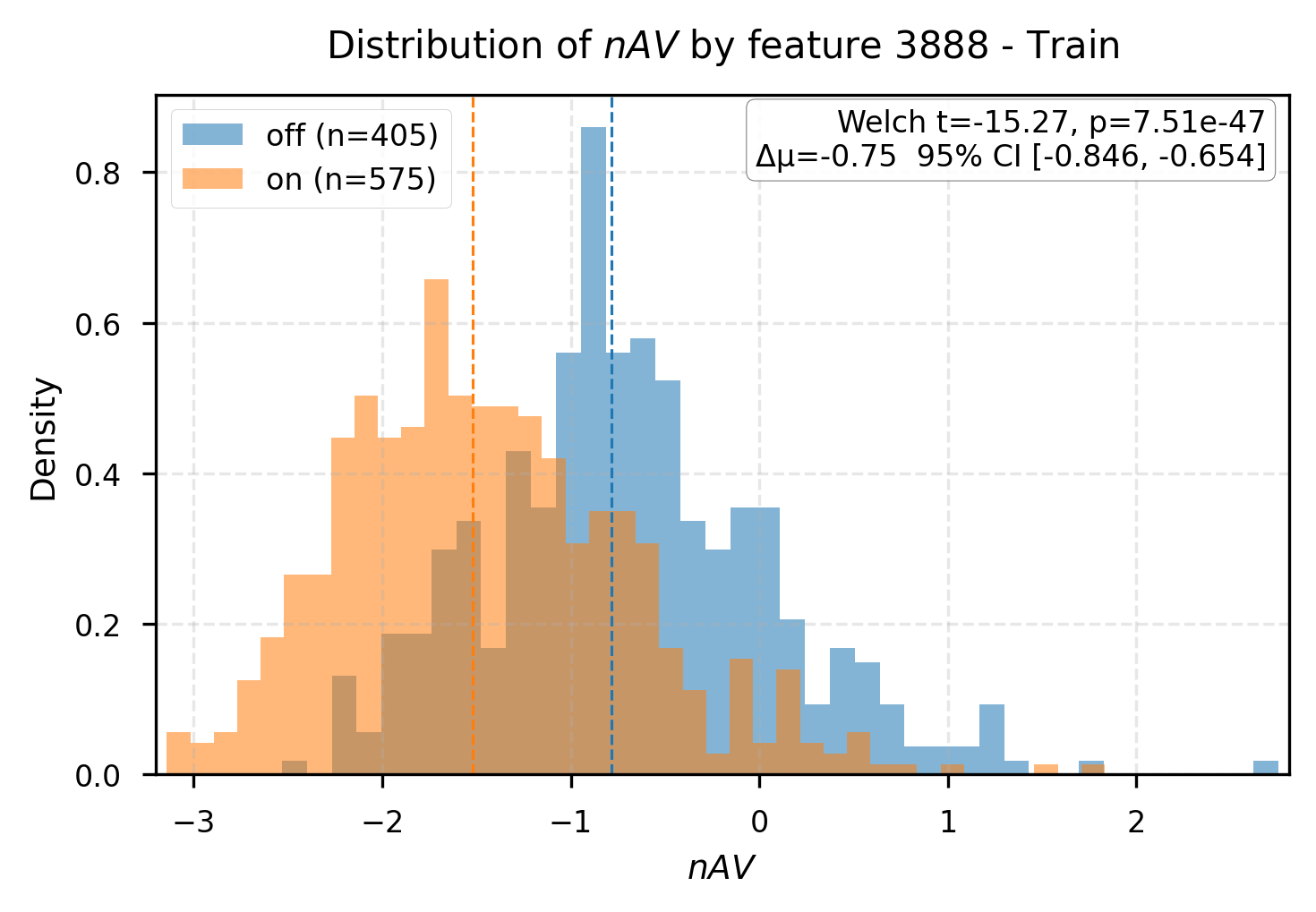}
    \includegraphics[width=0.49\linewidth]{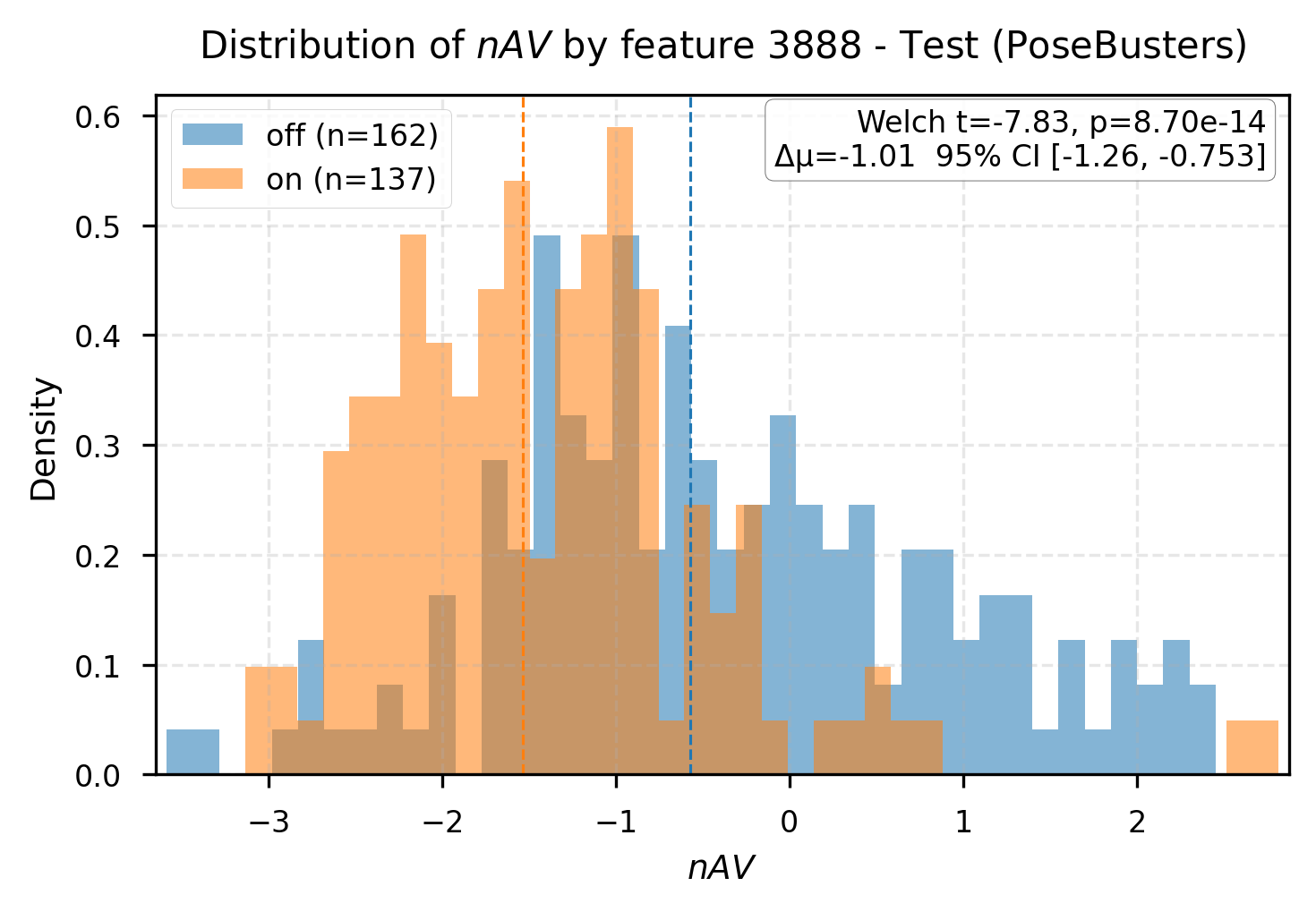}
    \caption{Group difference in Boltz-2 affinity values when splitting the dataset based on whether a feature activates or not. \textbf{Left:} feature 2299 from R3-L64, training set. \textbf{Right: }feature 3888 from R3-L33, training set. \textbf{Bottom: }feature 3888 from R3-L33, test set.}
    \label{fig:diff}
\end{figure}

In Section \ref{sec:hypo} we highlighted feature 2299 from R3-L64, that exhibits a strong group difference by activating on complexes with higher predicted affinity. In R3-L33 we found a feature representing the opposite, and activating only samples that on average have a lower predicted affinity. Group differences for these two features are presented in Figure \ref{fig:diff}.

\begin{figure}
    \centering
    \includegraphics[width=0.49\linewidth]{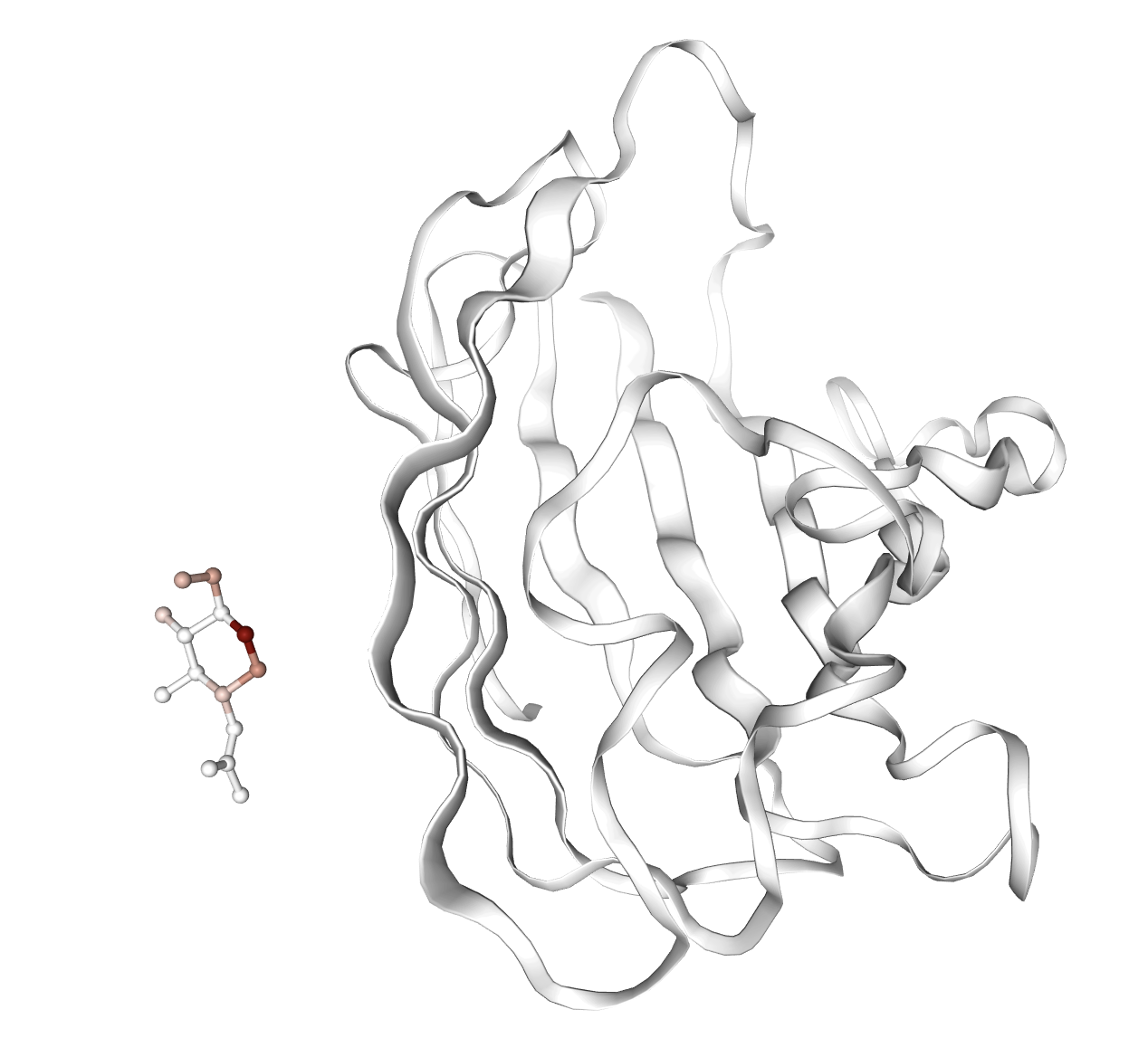}
    \includegraphics[width=0.49\linewidth]{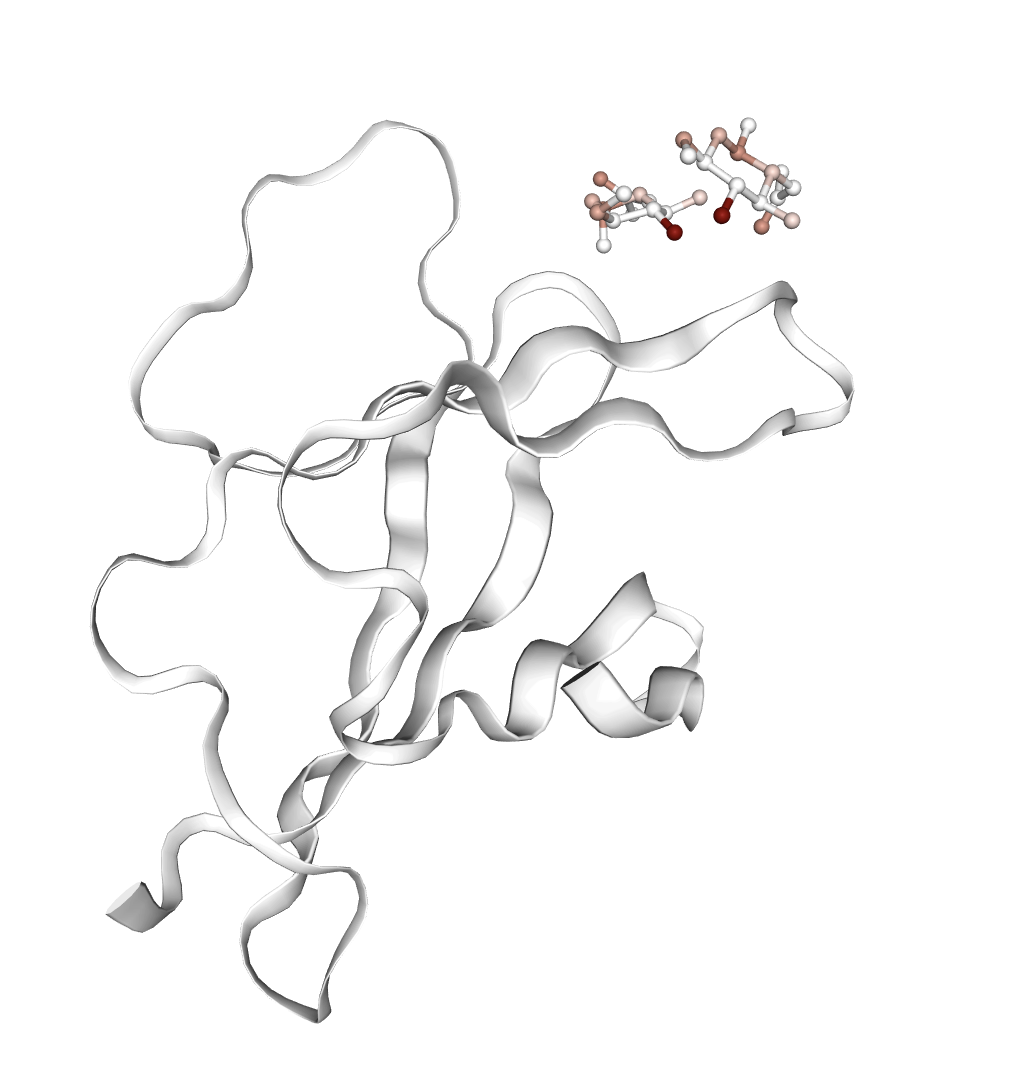}
    \includegraphics[width=0.5\linewidth]{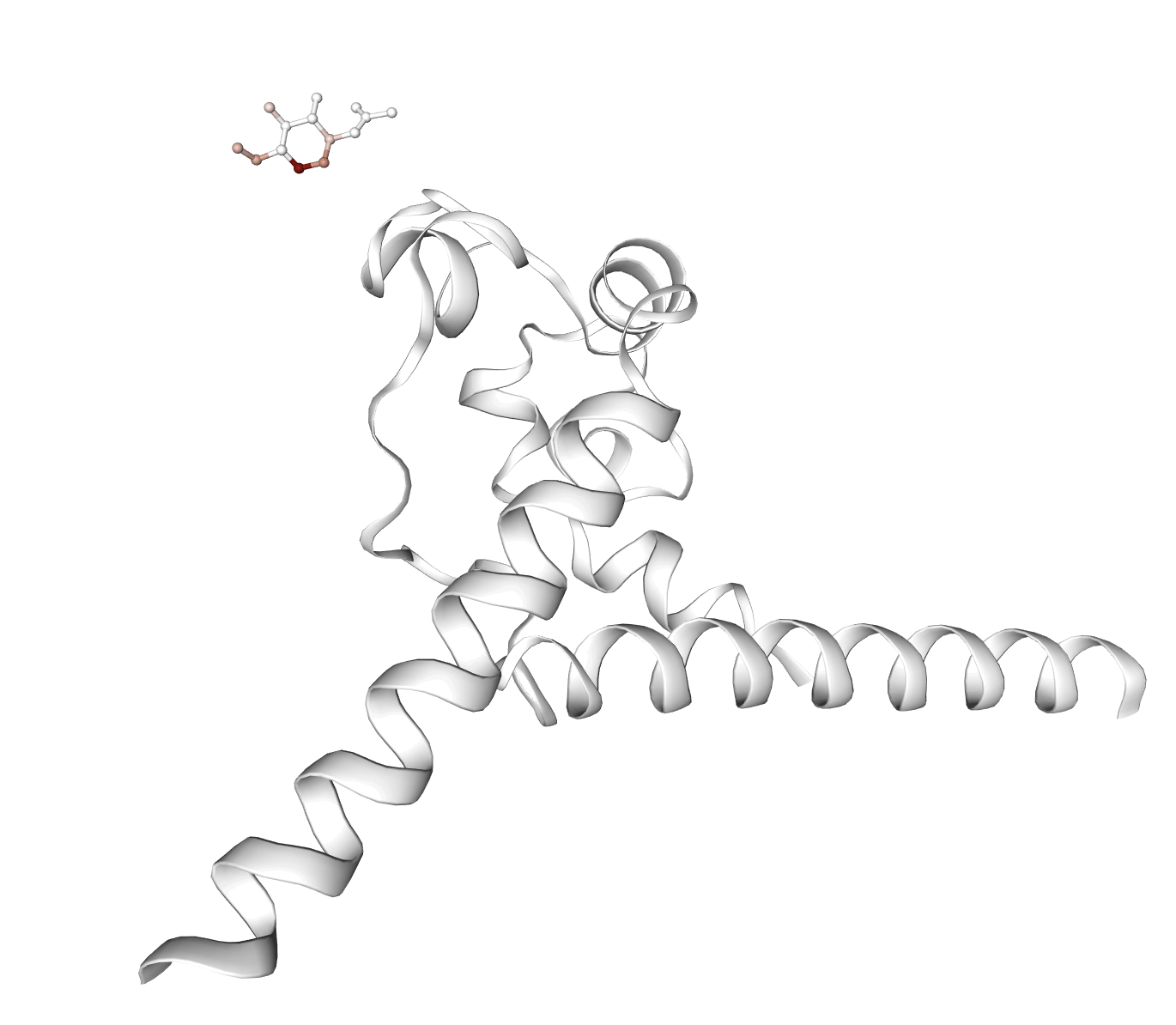}
    \includegraphics[width=0.8\linewidth]{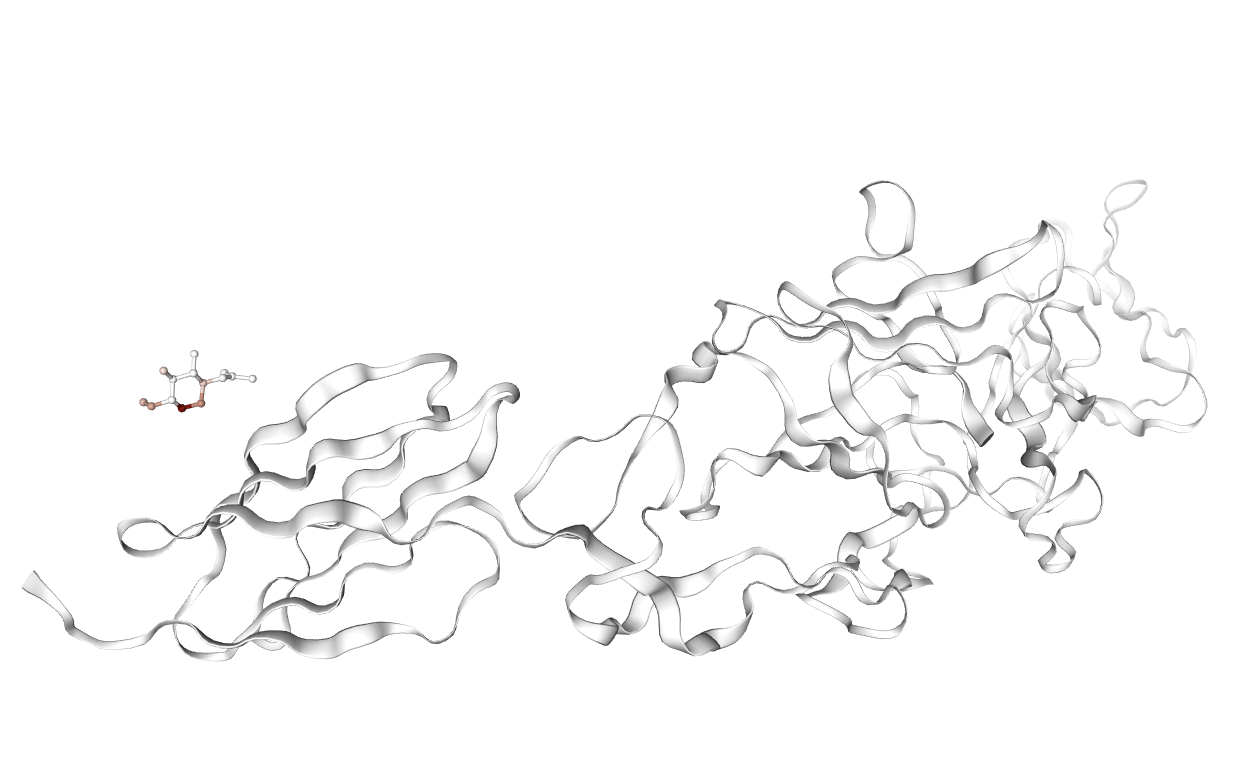}
    \caption{Examples where feature 3888 is activated.}
    \label{fig:3888}
\end{figure}

In Figure \ref{fig:3888} we display values of feature 3888 overlaid on ligands where it is activated. In these examples, the predicted complexes do not appear to present a plausible docked pose: the small molecule is displaced from the protein interface and does not form stable contacts.

\end{document}